\documentclass{article} %{svjour3}

\usepackage[margin=1in]{geometry}
\usepackage{cite}

\usepackage[pdftex]{graphicx}

\usepackage{color}
\definecolor{MyDarkRed}{rgb}{0.5,0,0.1}
\definecolor{MyDarkBlue}{rgb}{0.1,0.1,0.5}
\definecolor{MyDarkGreen}{rgb}{0.1,0.5,0.1}
\definecolor{MyRed}{rgb}{1.0,0,0}
\definecolor{MyBlue}{rgb}{0,0,1.0}
\definecolor{MyGreen}{rgb}{0,0.8,0}
\definecolor{lightgray}{rgb}{0.96,0.96,0.96}
\definecolor{darkgray}{rgb}{0.4,0.4,0.4}
\usepackage[pdftex, colorlinks=true, urlcolor=MyDarkBlue, citecolor=MyDarkRed, linkcolor=MyDarkGreen]{hyperref}

\usepackage{amsthm}
\usepackage{amsmath}
\usepackage{amsfonts}
\usepackage{amssymb}
\usepackage{upgreek}
\usepackage[dvipsnames]{xcolor}

\newtheorem{definition}{Definition}

\usepackage{algorithm}
\usepackage{algorithmicx}
\usepackage{algpseudocode}
\usepackage{array}

\usepackage[caption=false,font=footnotesize]{subfig}

\usepackage{color}
\usepackage[normalem]{ulem}

\newcommand{\changed}[1]{{#1}}

\begin{document}
\title{
Landmark-based Distributed Topological Mapping and Navigation in GPS-denied Urban Environments Using Teams of Low-cost Robots
}

%\titlerunning{Landmark-based Topological Mapping and Navigation}

%
%
%

\author{Mohammad Saleh~Teymouri \and Subhrajit~Bhattacharya% <-this % stops a space
\thanks{MS. Teymouri and S. Bhattacharya are with the Department
of Mechanical Engineering and Mechanics, Lehigh University, Bethlehem,
PA, 18015 USA e-mail: mot317@lehigh.edu, sub216@lehigh.edu (see https://www.lehigh.edu/~sub216/)}% <-this % stops a space
%\thanks{Manuscript received **, 2019; revised **, 2019.}
%\footnote{MS. Teymouri and S. Bhattacharya are with the %Department \at
%	Department of Mechanical Engineering and Mechanics, Lehigh University, Bethlehem, PA, 18015 USA. \\
%	%	Tel.: +123-45-678910\\
%	%	Fax: +123-45-678910\\
%	{\texttt{mot317@lehigh.edu, sub216@lehigh.edu}}           %  \\
%	%             \emph{Present address:} of F. Author  %  if needed
%	%	\and
%	%	S. Author \at
%	%	second address
%}
}


%\maketitle

\maketitle

\begin{abstract}
\changed{In this paper, we address the problem of autonomous multi-robot mapping, exploration and navigation in unknown, GPS-denied indoor or urban environments using a swarm of robots equipped with directional sensors with
limited sensing capabilities and limited computational resources.
The robots have no a priori knowledge of the environment and need to rapidly explore and construct a map in a distributed manner using existing landmarks, the presence of which can be detected using onboard senors, although little to no metric information (distance or bearing to the landmarks) is available.
In order to correctly and effectively achieve this, the presence of a necessary density/distribution of landmarks is ensured by design of the urban/indoor environment.
%
%This is achieved by using a topological representation of the environment called a \emph{Landmark complex}, which the swarm of robots build in a distributed manner through a balance between exploration and exploitation. 
%The robots are able to effectively use the Landmark complex for navigation in the environment.
%
We thus address this problem in two phases: 1) During the design/construction of the urban/indoor environment we can ensure that sufficient landmarks are placed within the environment. To that end we develop a \emph{filtration}-based approach for designing strategic placement of landmarks in an environment. 2) We develop a distributed algorithm using which a team of robots, with no a priori knowledge of the environment, can explore such an environment, construct a topological map requiring no metric/distance information, and use that map to navigate within the environment.
%This is achieved by using a topological representation of the environment called a \emph{Landmark complex}, which the robots build in a distributed manner through a balance between exploration and exploitation. 
%
%
This is achieved using a topological representation of the environment (called a \emph{Landmark Complex}), instead of constructing a complete metric/pixel map.
The representation is built by the robot as well as used by them for navigation through a balance between exploration and exploitation.
We use tools from homology theory for identifying ``\emph{holes}'' in the coverage/exploration of the unknown environment and hence guiding the robots towards achieving a complete exploration and mapping of the environment.
%
%We assume that each robot is equipped with a directional sensor with limited range of sensing
%that can detect landmarks within the sensor footprint, but cannot measure the bearing or the distance to the landmarks.
Our simulation results demonstrates the effectiveness of the proposed metric-free topological (simplicial complex) representation in achieving exploration, localization and navigation within the environment.
%
%Adopting concepts like \emph{Simplicial Complexes} and \emph{Landmark Complexes} facilitates topological modeling and algebraic hole-detection problems of such
%environments. A landmark complex is an abstract simplicial complex constructed by using identifiable landmarks in the environment. By taking benefits from these techniques, we are able to cumulatively
%construct the landmark complex which is the topological representation (metric-free model) of the environment. 
%
%First we propose an algorithm to place the minimum number of landmarks needed
%for exploration problem. Afterwards, we consider the exploration problem by submitting a multi-thread algorithm for swarm of decentralized robots with different modes of walk. And finally, we provide an
%algebraic approach for fixing the landmark complex coverage by using the homology theory.
}
%\keywords{Simultaneous Localization and Mapping (\emph{SLAM}), Swarm of Robots, Directional Sensors, Simplicial Complex, Landmark Complex, Coverage, Homology Theory.}
%Topological modeling and algebraic hole-detection of an unknown, GPS-denied environment are addressed by using swarm of decentralized robots with limited resources and directional sensors. Adopting
%concepts like ``Simplicial Complexes'' and ``Landmark Complexes'' facilitated the multi-thread implementation in C++ due to the optimum amount of information that need to be shared with other robots, by
%creating a metric-free model of the environment. Hence, in this paper, three algorithms have been presented to this end. First, an optimum mean of placing landmarks in the environment is proposed,
%afterwards, an algorithm is suggested for the decentralized multi robot exploration with different given motion planning, and finally, an algebraic approach for fixing the simplicial complex coverage has
%been presented. 

\end{abstract}

%\begin{IEEEkeywords}
%Simultaneous Localization and Mapping (\emph{SLAM}), Swarm of Robots, Directional Sensors, Simplicial Complex, Landmark Complex, Coverage, Homology Theory.
%\end{IEEEkeywords}

% \fi
%\IEEEpeerreviewmaketitle

\section{Introduction}

\subsection{Motivation}

Consider an unknown, GPS-denied urban/indoor environment in which we send out a large, fast-moving swarm of resource-constrained robots with extremely limited sensing capabilities and limited communication bandwidth, and with no a priori knowledge of the environment.
Precise range or bearing measurement to the landmarks in the environment may not be available (for example, in an urban environment such landmarks may be wireless routers or 5G antennae identified by their MAC ids, allowing simple wireless/5G receivers to detect only their presence, but not their precise range).
For communication and distributed exploration and mapping, each robot can potentially transmit only tens of bytes of data at a time.

An urban/indoor environment can be constructed/designed to aid such teams of robot to effectively explore, map, localize and navigate in.
This is relevant in context of search and rescue type operations in such environments where facilities are constructed to aid such operations when necessary.
While the swarm of robots itself may not have a blueprint of the environment, they can rely on the structured placement of reliable landmarks to aid with the process of exploration, mapping and navigation.

Without global localization or coordinate charts, our objective is to attain this in a metric-free and coordinate-free manner that does not rely on precise distance or bearing measurements, is fast, robust to errors, and does not require extensive filtering or post-processing in order to compensate for sensing and actuation noise.
%
%Without global localization or coordinate charts, our objective is to attain this in a \highlightContent{metric-free and coordinate-free manner that does not rely on precise distance or bearing measurements, is fast, is robust to errors, and does not require extensive filtering or post-processing in order to compensate for sensing and actuation noise}.
% 
In this paper we propose the use of \emph{simplicial complexes} as the metric-free, coordinate-free topological representations of the environment.
% Under certain assumptions on the density of landmarks in the environment and 
In particular, the robot swarm constructs an abstract simplicial complex representation, known as the \emph{landmark complex}, using the landmarks detected by them as they navigate through the environment. This is a low-fidelity but correct (\emph{homotopically equivalent}) topological representation of the free space (under appropriate assumptions on the density of landmarks and the coverage attained by the swarm).

\subsection{Related Works}

While \emph{GPS} is broadly available to users around the world for localization and navigation, and so are maps of urban environments, a reliance on such information is not practical in many contexts. 
For example, GPS or maps may not be reliably available inside buildings with thick concrete walls. 
Since they require complex infrastructure to operate, GPS or a global map database may not be available for underground or extraterrestrial (such as future Lunar or Martian) colonies.

Most state-of-the-art methods for construction of maps of unknown environments without localization fall under the SLAM literature and require precise metric information (such as range or bearing measurements), rely on relatively precise odometry measurements, and in order to build a complete map, require extensive post-processing for correcting accumulated errors~\cite{castellanos1999spmap, choset2001topological, montemerlo2002fastslam, durrant2006simultaneous, davison2007monoslam, angeli2008fast}.
% Such processes of simultaneous localization and mapping (SLAM)
% Traditional discrete representations of environments have been 
% The representations hence constructed are usually 
Such methods usually construct a grid-based coordinate representation, making the amount of information that need to be shared between robots 
% in such traditional methods is 
extremely large, and a precise transformation between the different robots' grid maps difficult to compute~\cite{thrun2005multi, howard2006multi}.
State-of-the-art visual odometry based localization, mapping and navigation require significant sensing capabilities (such as stereo or monocular cameras)~\cite{engel2015large,munguia2007monocular,quan2019tightly,liu2019accurate,10.1007-978-3-319-10605-2_54} in order to determine relatively precise metric information about detected features and landmarks in the environment, and focuses on precise pose estimation of the robots.
Various implementations of {SLAM} can be found in different industries nowadays such as
self-driving autonomous vehicles \cite{Autonomous_vehicle_1, Autonomous_vehicle_2,forster2014svo,mur2015orb} and consumer robot vacuum cleaners~\cite{vacuum_clear_robot}. These state-of-the-art methods need meticulous metric measurement tools like range measuring sensors, relatively precise odometry measurements, and expensive cameras.

We, on the other hand, use topological methods to address the problem where multiple robots with only on-board limited-range sensors can detect presence of landmarks within their respective sensing disks (the binary information of whether a landmark is present or not), but no distance measurements to the landmarks are used. 
We consider a multi-robot setup in which a large number of robots need to cooperatively build the topological representation of the environment. Without knowing its own location, nor the locations of other robots in the environment (globally or relative to itself), the robots only communicate to other robots the identity of the landmarks that they observe -- an extremely small amount of data -- which allows the swarm to build the map collectively and in a distributed fashion.

While our method does not intend to match or compete with the metric precision of high-fidelity state-of-the-art SLAM techniques, the strength of our method lies in the use of extremely low-fidelity and inexpensive sensing and computational capabilities that allow the robots to perform mapping, localization and navigation tasks without requiring such precision.
We consider directional sensors on robots such that the binary information of existence of a landmark can be detected by the robot in its sensor footprint. This makes the configuration space in which the robots need to construct the simplicial complex representation (``\emph{landmark complex}'') a subset of $SE(2)$. Although we consider constrained resources for robots, {landmarks} are assumed to be available in necessary density for the topological exploration and mapping of the environment. The robots however cannot measure the bearing or the distance to the landmarks. The only
information a robot uses for local control, is whether a detected landmarks is to its left or to its right side.

% The problem in general involves two steps:
% The two steps that we propose in doing so are 
% Key to solving a localization and mapping problem is to ensure a set of features or landmarks

% The environment being devoid of features, the first task at hand is to establish landmarks in the environment, using which robots can create a coordinate-free topological representation of the environment for the purpose of navigation.
% 
% The method we propose is purely topological in nature which uses a simplicial complex as the representation of the environment.  Specifically, we consider a \emph{landmark complex,} a simplicial complex constructing by the observation of landmarks ~\cite{ghrist2012topological}.

While \emph{landmark complex}~\cite{ghrist2012topological,ICRA:18:landmark},
and more generally, simplicial complexes~\cite{dirafzoon_lobaton_2013,jasons_paper} have been used for topological representation of environments,
most of the existing work in this field has been for robots with disk-shaped sensor footprints in a planar domain. 
Furthermore, current literature does not explicitly address the problem of planning navigation of robots 
% to gather information that would allow construction of the complete landmark complex.
using the landmark complex, either to gather information that would allow construction of the complex itself or to execute tasks. %such as coverage, transportation or surveillance.
%Furthermore, there has been no consideration for uncertainty in the detection of presence of landmarks, uncertainty in properly identifying them, or absence of landmarks altogether.
% 
In this paper we address these practical issues concerning the construction and exploitation of a landmark complex representation.
We also design navigation algorithms for the robots constructing the landmark complex representation through exploration of the environment as well as for robots exploiting the landmark complex for goal-directed navigation. %, transportation, coverage and surveillance.

\subsection{Contribution and Organization}

The main contributions of the paper can be broadly classified into three parts:
1) The development of a filtration-based algorithm for determining placement of landmarks in an urban/indoor environment during the design/construction of the environment;
2) Design a set of controllers that would allow a team of robots to perform exploration of the environment for constructing the Landmark complex representation;
3) Design a set of controllers that would allow the team of robots to exploit the partially or fully constructed Landmark complex to perform informed exploration or navigation within the environment.

The paper is organized as follows:
In Section~\ref{sec:preliminaries} we introduce the basic definitions of \emph{Simplicial Complex} and \emph{Landmark Complex}.
In Section~\ref{sec:landmark-placement} we describe the filtration-based algorithm for strategic placement of landmarks in an urban/indoor environment during its design and construction.
Section~\ref{sec:exploration-navigation} provides a set of control algorithms for the robots in the swarm which they can use to explore an unknown environment for constructing the Landmark complex and exploit the (fully or partially) constructed Landmark complex for further exploration and navigation. In this section we also use tools from homology theory for detecting `\emph{holes}'' in the exploration and hence filling them.
In Section~\ref{sec:results} we provide detailed simulation results and evaluation of the proposed algorithms.

\section{Preliminaries} \label{sec:preliminaries}
In this section we provide brief definitions of \emph{Simplicial Complex} and \emph{Landmark Complex}.

\subsection{Simplicial Complexes and Simplices}
\begin{definition}[An abstract Simplicial Complex~\cite{Hatcher:AlgTop}]
An abstract simplicial complex, $\mathcal{K}$, constructed over a set $V$ (the \emph{vertex set}) is a collection of sets $C_n, ~n=0,1,2,\cdots$, such that:

\textbf{\emph{i.}}
An element in $C_n, ~n\geq 0$ is a subset of $V$ and has cardinality $n+1$ (\emph{i.e.} for all $\sigma \in C_n$, $\sigma \subseteq V, |\sigma| = n+1$. $\sigma$ is called a ``\emph{$n$-simplex}''), and,

\textbf{\emph{ii.}}
If $\sigma \in C_n, ~n\geq 1$, then $\sigma \!-\! v \in C_{n-1}, ~\forall v \in \sigma$. Such a $(n$-$1)$-simplex, $\sigma \!-\! v$, is called a ``\emph{face}'' of the simplex $\sigma$.
The simplical complex is the collection $\mathcal{K} = \{C_0, C_1, C_2, \cdots\}$.
\end{definition}

\begin{figure}[h]
\centering
\includegraphics[width=3in]{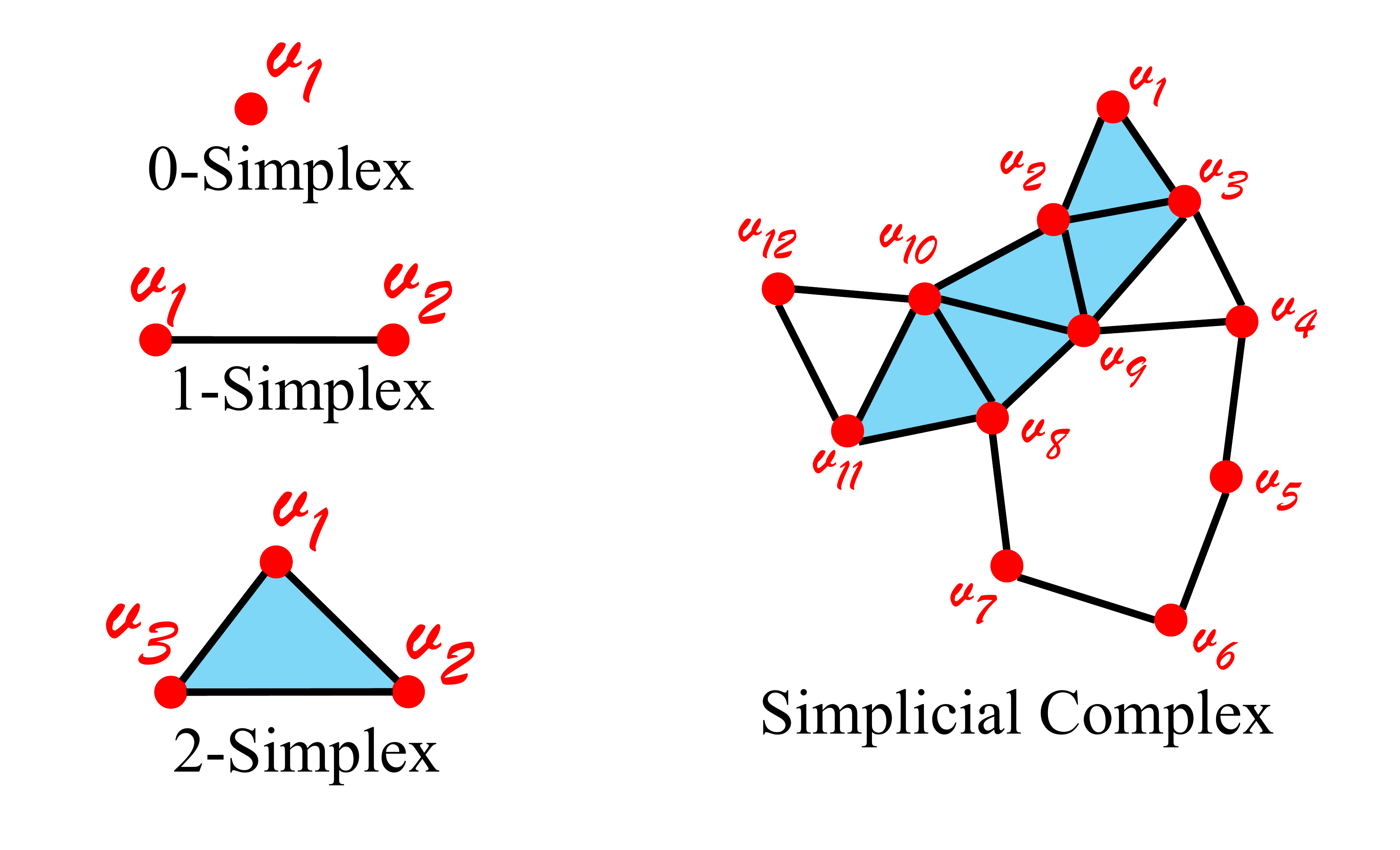}
\caption{An abstract simplicial complex consists of simplices of different dimensions. In this figure, the simplicial complex on the right, constructed from $0$-simplices, $1$-simplices and
$2$-simplices, depicted individually on left.}
\label{simplicial complex}
\end{figure}

Moreover, in Figure \ref{simplicial complex} an example of simplicial complex is provided. While a graph constitutes of only two sets ($V$ and $E$), in this example a simplicial complex with three sets is
presented. In this figure, $\mathcal{K} = \{ C_0, C_1, C_2 \}$, where $C_0$ is the set of all $0$-simplices (\emph{vertex set}), $C_1$ is the set of all $1$-simplices (\emph{edge set}) and $C_2$ is the set
of all $2$-simplices, representing a triangle that connects three vertices. In this example, the $C_0$, $C_1$ and $C_2$ are as following:

\textbf{\emph{i.}} $C_0 = V = \{v_1, v_2, ..., v_{12}\}$.

\textbf{\emph{ii.}} $C_1 = E = \{ \{v_1, v_2\}, \{v_1, v_3\}, \{v_2, v_3\}, ..., \{v_{11}, v_{12}\} \}$.

\textbf{\emph{iii.}} $C_2 = \{ \{v_1, v_2, v_3\}, \{v_2, v_3, v_9\}, \{v_2, v_9, v_{10}\}, \{v_8, v_{10}, v_{11}\} \}$.

\subsection{Landmark Complex}

A Landmark Complex~\cite{ghrist2012topological,ICRA:18:landmark}, $\mathcal{X}$, is an abstract simplicial complex, composed of the set of landmarks observed by a robot navigating in its configuration space.
Assuming the environment has sufficient landmarks (the precise criteria is described in Section~\ref{sec:landmark-placement}),
a robot's task is to collect the information of the landmarks, and create an abstract
simplicial complex which is called Landmark Complex. Every time a robot observes a $n$-tuple of landmarks, it inserts the corresponding $(n-1)$-simplex constituting of the observed landmarks in
$C_{n-1}$ (along with all its faces and sub-faces in $C_i, i<n-1$). Furthermore, the information contained in the landmark complex can be interpreted to generate the map of the environment.

\begin{figure}
\centering
\includegraphics[width=3in]{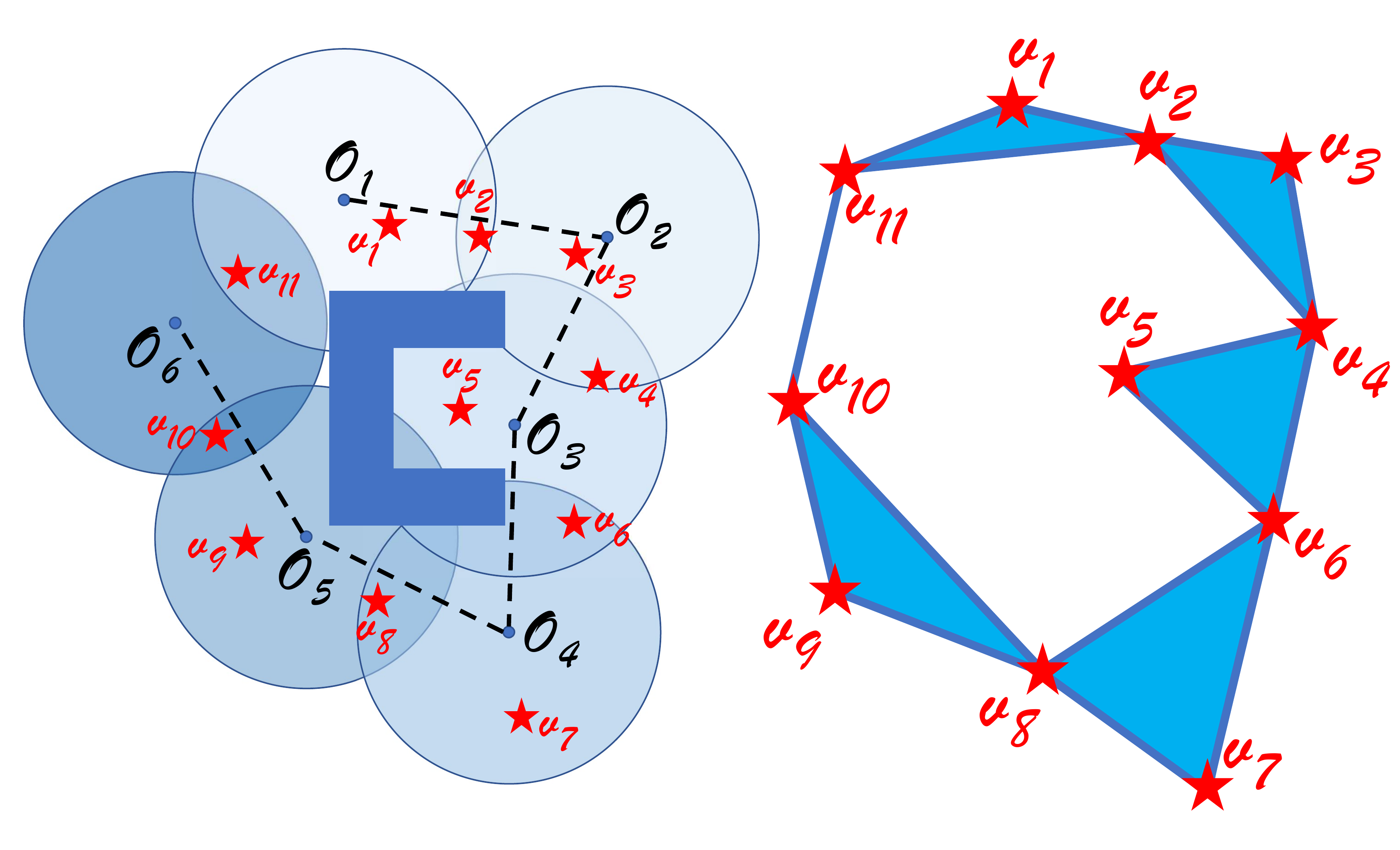}
\caption{On the left there is an environment with 11 landmarks depicted as stars, and the corresponding landmark complex constructed from 6 observations with omni-directional sensor is shown on the right.}
\label{landmark complex}
\end{figure}

Figure \ref{landmark complex} illustrates an example of creating a landmark complex by 6 observations with omni-directional sensor and use it to generate the topological map of the environment. In this example,
for the first observation point $O_1$, the robot observes the landmarks $\{v_2, v_3, v_4\}$ which actually is a $2$-simplex. But, whenever a $2$-simplex is observed, the connecting lines will also be
considered as a $1$-simplex, so in this case there exists $1$-simplex as following, $C_1 = \{\{v_2, v_3\}, \{v_2, v_4\}, \{v_3, v_4\}\}$ and as the same way the vertices are forming $0$-simplex as well.
This means, every time that we have an observation forming an $n$-simplex the lower dimension simplices will be inserted into the landmark complex. So, after 6 observations the created landmark
complex would be $\mathcal{X} = \{ C_0, C_1, C_2 \}$ where the $0$-simplex or the $Vertices$ set is equal to $C_0 = \{v_1, v_2, v_3, ..., v_{11}\}$, the $Edges$ set or $1$-simplex is
$C_1 = \{ \{v_1, v_2\}, \{v_2, v_3\}, \{v_2, v_4\}, \{v_3, v_4\}, ..., \{v_{10}, v_{11}\} \}$ and the $2$-simplex is
$C_2 = \{ \{v_2, v_3, v_4\}, \{v_4, v_5, v_6\}, \{v_6, v_7, v_8\},$\\$ \{v_8, v_9, v_{10}\}, \{v_{11}, v_1, v_2\} \}$.

The Landmark Complex $\mathcal{X}$ can be \emph{immersed} on a plane to create a visual representation of the topological map as shown in Figure \ref{landmark complex}. As visible in the image of $\mathcal{X}$, the landmark complex captures the obstacle in the environment in terms of a \emph{hole} in the complex.
%where the big hole in the image represents the existence of an obstacle in the environment.

\section{Landmark Placement Algorithm} \label{sec:landmark-placement}

The ability to plan and implement strategic landmark placement in the environment allows faster robot explorations and precise topological mapping of the environment.
For instance, in situations like search and rescue
missions, if a building is equipped with sufficient landmarks (beacons) during its construction, firefighters can use a swarm of distributed robots to explore the environment and locate survivors more effectively and safely. 
Moreover, in hazardous places where the environment is too harsh for humans to operate (such as in nuclear power plants), having strategically placed landmarks in 
the environment allows robots to perform maintenance operations autonomously.
This is made possible in urban or indoor environments, which, during their design and construction, can be built with the planned placement of landmarks.
In this section we describe an algorithm for strategically placing landmarks in an environment in order to attain such objectives.

\subsection{Domain of Visibility}
\begin{figure}[!t]
\centering
\subfloat[]{\includegraphics[width=2in]{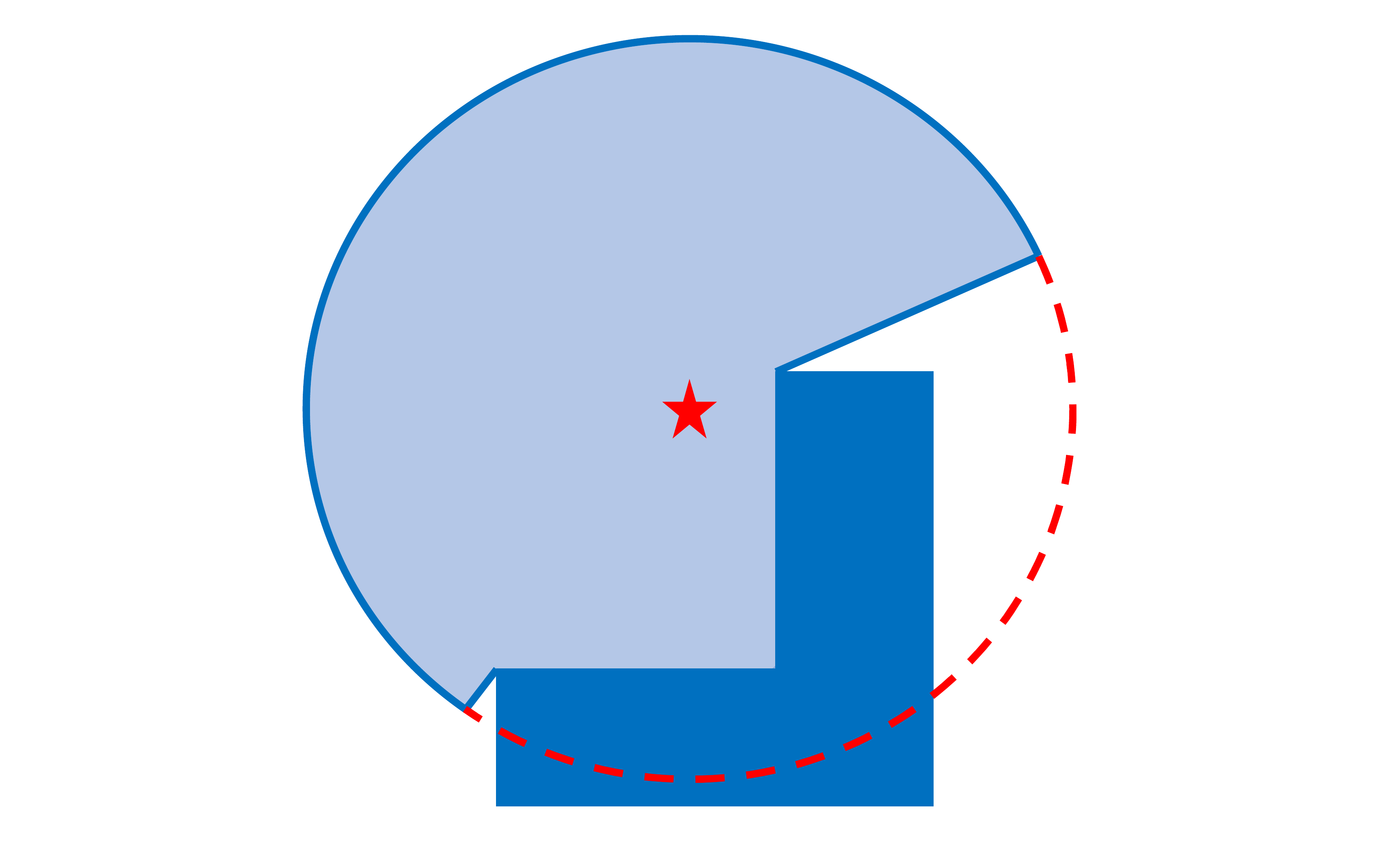}
\label{R2}}\hfil
\subfloat[]{\includegraphics[width=1.39in]{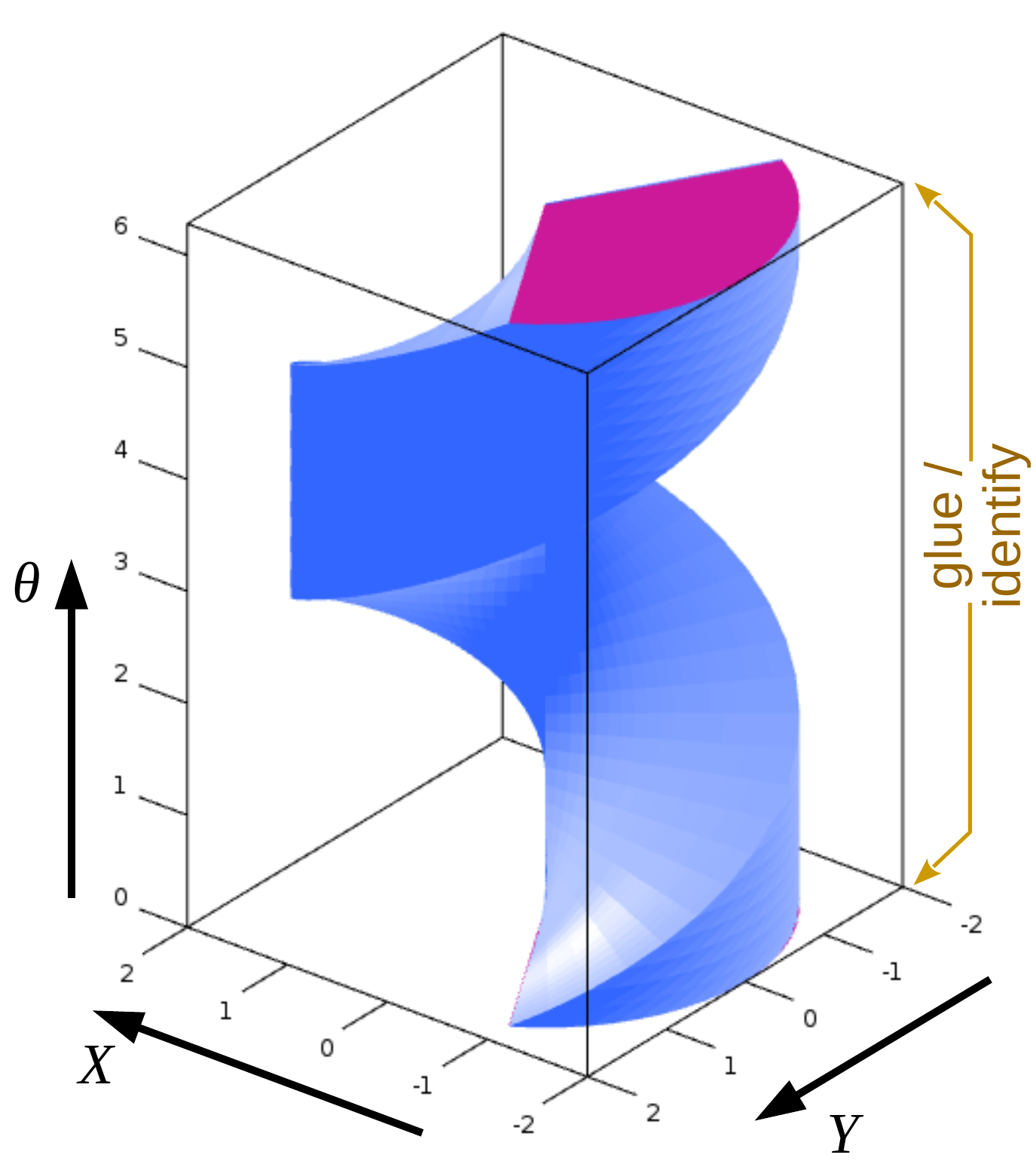}
\label{se2}}\hfil
\caption{\textbf{Domain of Visibility: }Figure (a) shows the domain of visibility for ${\mathbb{R}^2}$, and Figure (b) shows it for $SE(2)$.}
\label{domain of visibility}
\end{figure}

We define the \emph{workspace}, $\mathcal{W} = \mathbb{R}^2 - O$, to be the set of all possible landmark locations (where $O$ is the set of all obstacles in the environment) and the \emph{configuration space}
$\mathcal{C}$ which is the set of all robot's configurations. We also define the visibility domain of a landmark, $\mathcal{D}_{p, S} \subset \mathcal{C}$, to be the set of all robot configurations in which the landmark at
location $p \in \mathcal{W}$ will be visible to a robot using sensor footprint $S \subset \mathcal{W}$.
Figure \ref{R2} shows a situation where both $\mathcal{W}$ and $\mathcal{C}$ are subset of
$\mathbb{R}^2$. Therefore, In this figure, the sensor footprint is in shape of a disk. However, in this paper, robots with directional sensors (sector of a disk) are being considered, and accordingly the
configuration space $\mathcal{C}$ for is a subset of $SE(2)$. Figure \ref{se2} shows the visibility domain of a landmark in $SE(2)$. In this figure, the workspace $\mathcal{W}$ is still
$\mathbb{R}^2$ representing the possible landmark's positions.

\subsection{Nerve Lemma and Visibility Condition}
\begin{figure}
\centering
\subfloat[\u{C}ech complex of the visibility domains]{\includegraphics[width=2.5in]{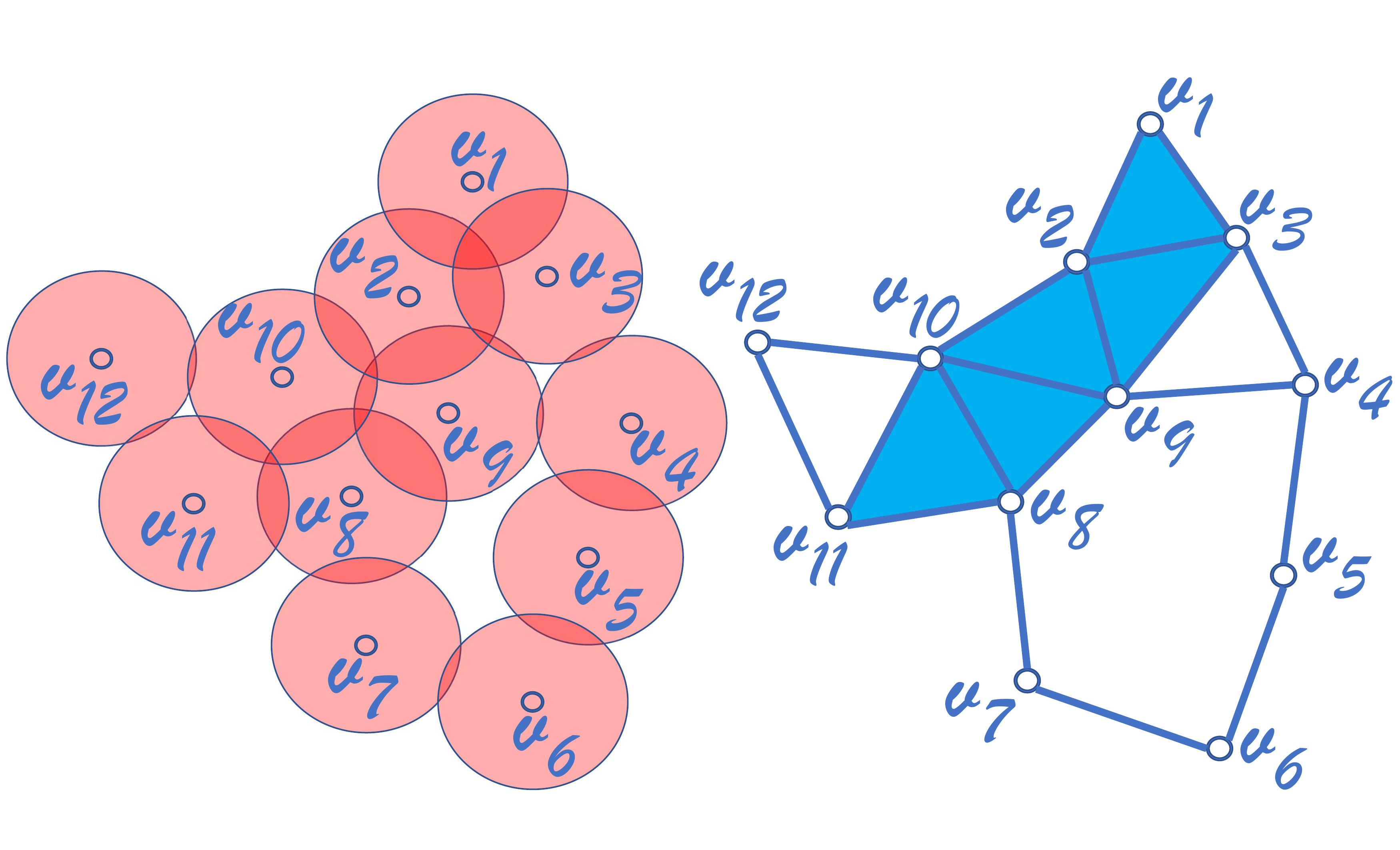}
\label{cech vs landmark complex (a)}}\hfil
\subfloat[Landmark complex constructed over 12 observations]{\includegraphics[width=2.5in]{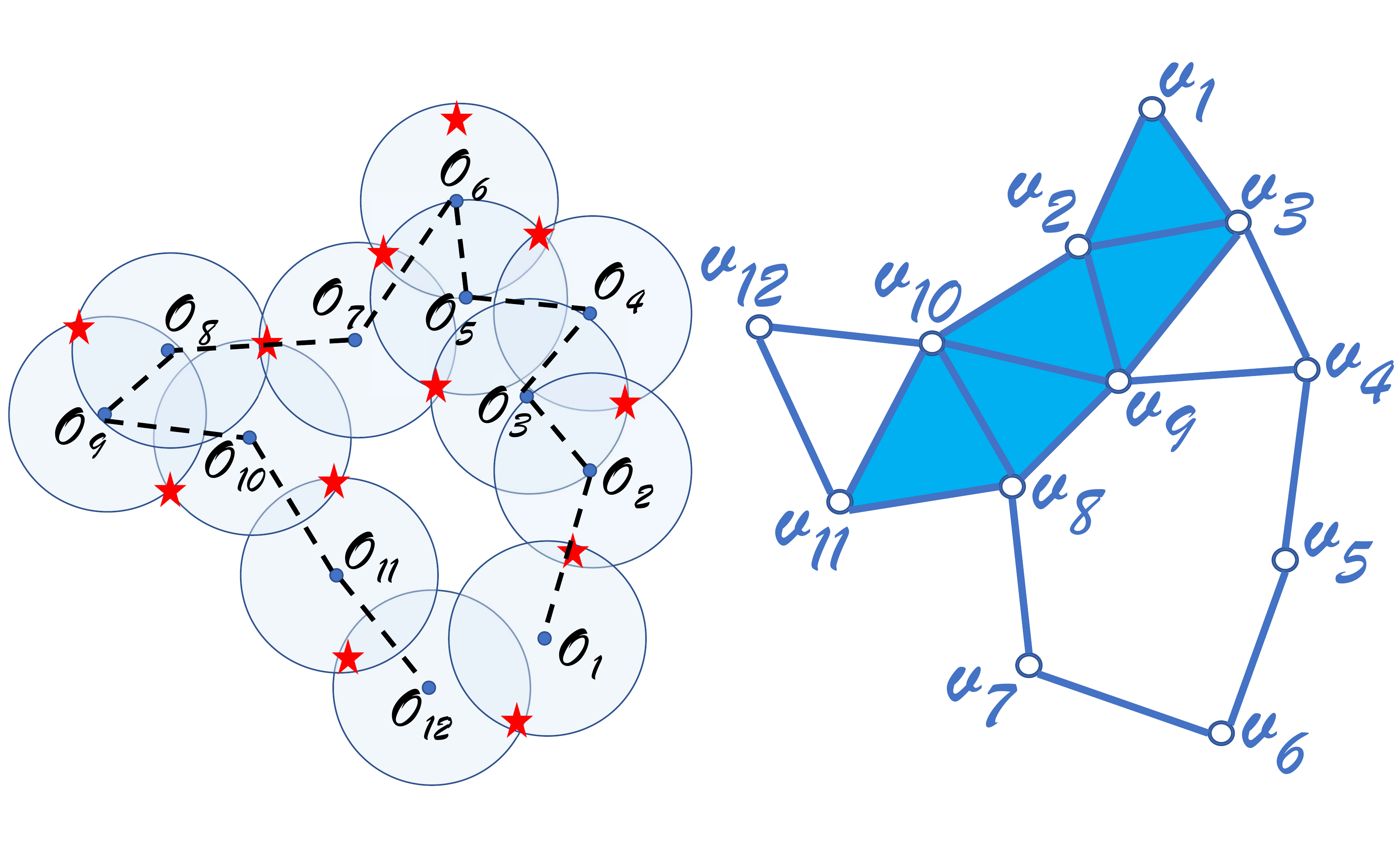}
\label{cech vs landmark complex (b)}}\hfil
\caption{\u{C}ech complex vs. Landmark Complex}
\label{cech vs landmark complex}
\end{figure}

The overall objective of this section is to establish the visibility and density condition for the presenting Landmark Placement Algorithm (LPA). First we show that the \u{C}ech complex~\cite{Hatcher:AlgTop},
$\mathcal{H}$, constructed by the visibility domain of the landmarks will be homotopy equivalent to the Landmark Complex $\mathcal{X}$ constructed during the robot explorations. According to the
definition of the \u{C}ech complex, whenever there is a $n$-way intersection of visibility domains, there would be a corresponding $(n$-$1)$-simplex in $\mathcal{H}$. Furthermore, by definition, a
visibility domain is a set of all configurations that a landmark is visible to a robot. Therefore, whenever there is a $(n-1)$-simplex in $\mathcal{H}$, it implies that if a robot have a configuration
in the intersection of those visibility domains, then all of the $n$ landmarks will be visible to the robot and there would be the corresponding simplex in $\mathcal{X}$ too. Therefore,
$\mathcal{H} \cong \mathcal{X}$. In Figure \ref{cech vs landmark complex}, an example is shown for more elaboration.

Suppose $\mathcal{P} = \{p_1, ..., p_n\}$ is the set of landmark positions, and $\mathcal{M} = \{\mathcal{D}_{p, S} \subset \mathcal{C} \mid p \in \mathcal{P}\}$ is the set of landmark visibility domains. Using the \emph{Nerve Theorem} from \emph{algebraic topology}~\cite{Hatcher:AlgTop}, if the open cover $\mathcal{M}$ of a topological space $\mathcal{C}$ is a \emph{good cover}, then the nerve of the cover $\mathcal{N(R)}$ is \emph{homotopy equivalent}
to the topological space $\mathcal{C}$ without missing any topological information of the space. By definition, an open cover $\mathcal{M}$ is considered a good cover when not only every nonempty sets of
$\mathcal{M}$, but every two and three way intersections are contractible. In other words, the \u{C}ech Complex constructed by the visibility domain of the landmarks $\mathcal{H}$ is homotopy equivalent to
the configuration space $\mathcal{C}$ ($\mathcal{H} \cong \mathcal{C}$) and previously it has been shown that $\mathcal{H} \cong \mathcal{X}$ consequently $\mathcal{X} \cong \mathcal{C}$. Therefore, during
the landmark placement phase, there should be at least one landmark visible to the robot for every configuration in the configuration space to obtain the metric-free map of the environment.

\subsection{Landmark Placement Algorithm using Filtration over Sensor Footprints}
In this section an iterative algorithm is presented to place the landmarks such that the \u{C}ech Complex $\mathcal{H}$ covers the entire configuration space $\mathcal{C}$. To this end, we use a filtration over sensor footprints -- if $S^s$ is the true sensor footprint of a robot's sensor, we define $S^1 \supset S^2 \supset ... \supset S^s$ to be a sequence of sensor footprints such that
$S = \{S^t \subset \mathcal{C} \mid 1 \leq t \leq s\}$, where $S^t$ is the sensor footprint to be used at $t^{th}$ iteration.

The overall algorithm is to place a set of new landmarks in the \emph{uncovered} regions of the environment at the $t^\text{th}$ iteration, considering $S^t$ to be the sensor footprint at that iteration. Starting with a relatively large sensor footprint, $S^1$, we gradually decrease the size of the footprint until we attain $S^s$, while placing landmarks in the uncovered regions in every iteration (Figure~\ref{Landmark Placement Algorithm} illustrates the idea for disk-shaped sensor footprints).
More details of the algorithm are provided below.

Denote the position of the $i^{th}$ landmarks by $p_i$ and $\mathcal{P}^t = \{p_i \in \mathcal{W} \mid 1 \leq i \leq n\}$ is the set of all landmark positions at iteration $t$, where $\mathcal{W}$
is the workspace. Furthermore, we define $\mathcal{D}_{\mathcal{P}^t, {S}^t} = \bigcup\limits_{p \in \mathcal{P}^t} \mathcal{D}_{p, S^t}$ as the union of all landmarks' visibility domains, such
that $\mathcal{D}_{\mathcal{P}^t, {S}^t} \subset \mathcal{C}$. Suppose $\mathcal{U}^t$ is the complement of the covered space $\mathcal{D}_{\mathcal{P}^t, {S}^t}$ at $t^{th}$ iteration where
$\mathcal{U}^t$ = $\mathcal{C}$ - $\mathcal{D}_{\mathcal{P}^t, {S}^t}$ and $\mathcal{U}^t \subset \mathcal{C}$. At every iteration, the Landmark Placement Algorithm $(LPA)$ detects
$U_i^t,~i = 1, 2, ..., m$, the $i^{th}$ connected component of the uncovered area $\mathcal{U}^t$ at $t$ such that $\mathcal{U}^t$ = $\bigcup\limits_{i = 1}^{m} U_i^t$, and suppose $\bar{U}^t$ is the
set of all connected components $\bar{U}^t = \{U^t_1, U^t_2, U^t_3, ..., U^t_m\}$.
In order to cover $U_j^t$ for $j$ = $1, 2, ..., m$, $LPA$ places a landmark at $p_j$ such that $\mathcal{D}_{p_j, S^t}$ covers most of the $U_j^t$ and inserts $p_j$ into the $\mathcal{P}^t$. Afterwards
$LPA$ will compute the new $\mathcal{U}^t$ with newly placed landmarks and continues the same procedure until $\mathcal{U}^t = \emptyset$. In the next iteration, $LPA$ goes to the next sequence of the
sensor footprints, $S^{t+1}$ such that $S^{t+1} \subset S^t$ and $\mathcal{D}_{\mathcal{P}^t, S^{t+1}} \subset \mathcal{D}_{\mathcal{P}^t, S^t}$. Since the net cover of
$\mathcal{D}_{\mathcal{P}^t, S^{t+1}}$ is smaller than the cover domain with $S^t$, new connected components of uncovered area may open up in the configuration space $\mathcal{C}$. Therefore, new landmark
positions need to be populated in $\mathcal{P}^{t+1}$. We start with $\mathcal{P}^{t+1} = \mathcal{P}^t$ and then the $LPA$ process restarts by identifying the connected components in
$\mathcal{U}^{t+1} = \mathcal{C}$ - $\mathcal{D}_{\mathcal{P}^{t+1}, S^{t+1}}$.

\begin{figure}[!t]
\centering
\includegraphics[width=2in]{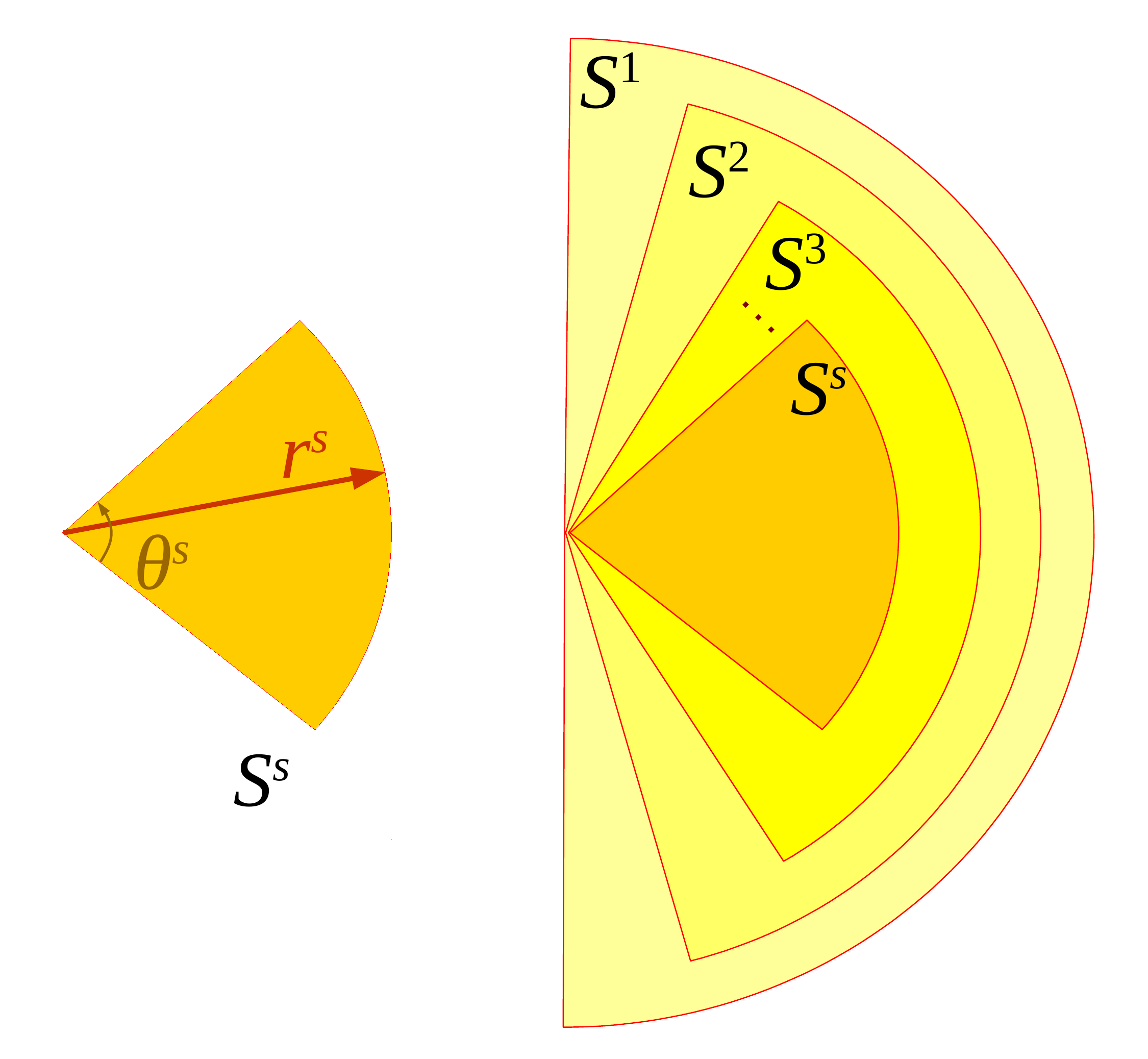}
\caption{Sensor footprint sequence for directional sensors ($SE(2)$).}
\label{sensor footprint sequence}
\end{figure}

At the very first iteration, the $LPA$ starts with a relatively large visibility domain, $\mathcal{D}_{p_1, S^1}$, placed at the center of the workspace $\mathcal{W}$ such that in an obstacle-free
environment $\mathcal{D}_{p_1, S^1} \cong \mathcal{C}$. In circumstances that the sensor footprint sequence is consisted of concentric disks ($\mathcal{W} = \mathcal{C} \subset \mathbb{R}^2$), $S^1$ is a
big circle where its radius is larger than the diameter of the environment. During the filtration, the radius of $S^1$ gets decreased until the target sensor footprint $S^s$ is reached. On the other hand,
in case of directional sensor footprints where $\mathcal{C} \subset SE(2)$, $LPA$ starts with a large half disk sensor footprint, $S^1$, where its radius and the angle of the circular sector reduces during
the filtration until it reaches $S^s$. Figure \ref{sensor footprint sequence} shows a sequence of sensor footprints in $SE(2)$.

\begin{figure}[t]
\centering
\includegraphics[width=3.2in]{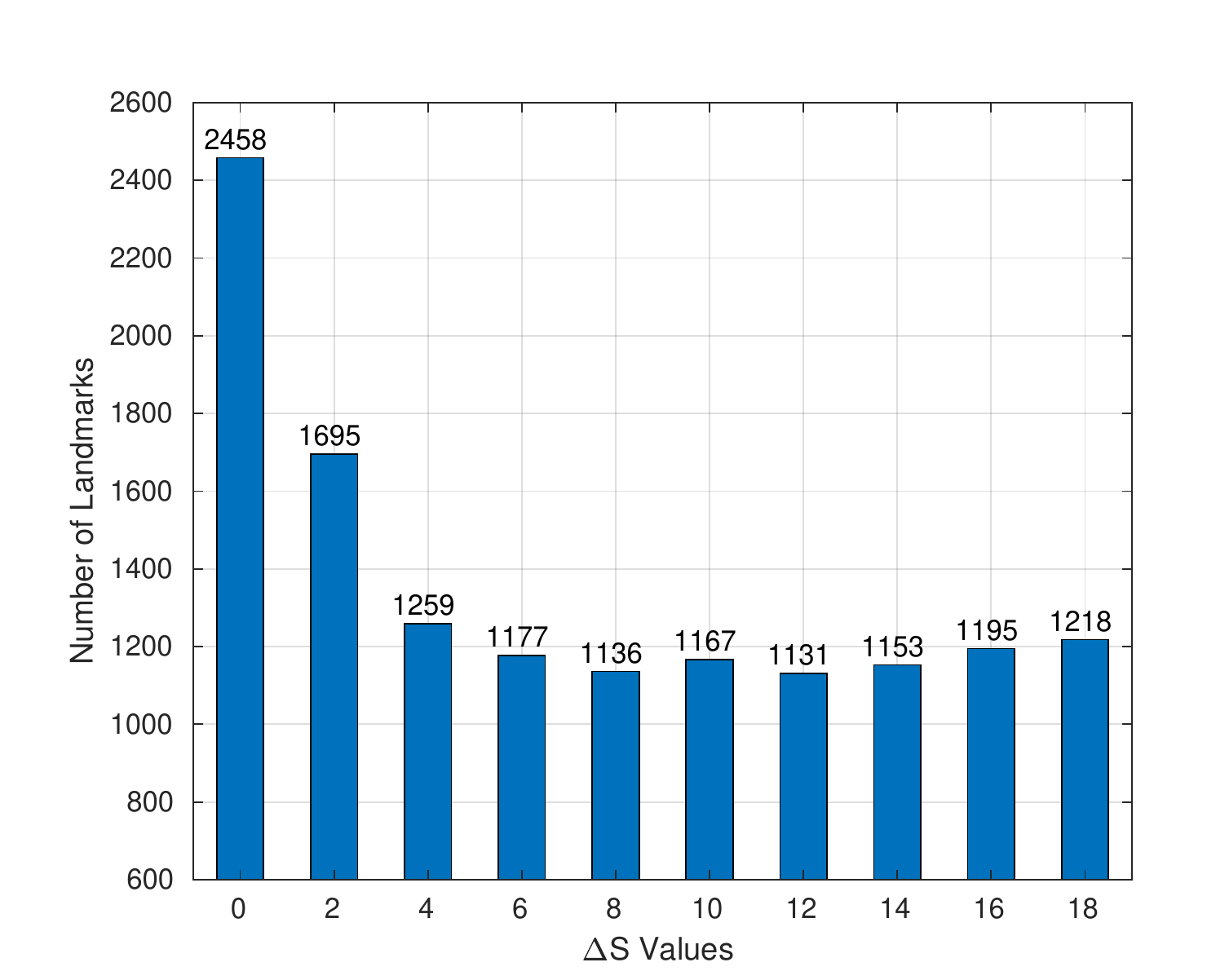}
\caption{Number of landmarks for different $\Delta S$ values.}
\label{LPA_Minimize}
\end{figure}

\begin{figure*}[!t]
\centering
\subfloat[]{
\begin{minipage}{\linewidth}
\includegraphics[width=0.24\linewidth]{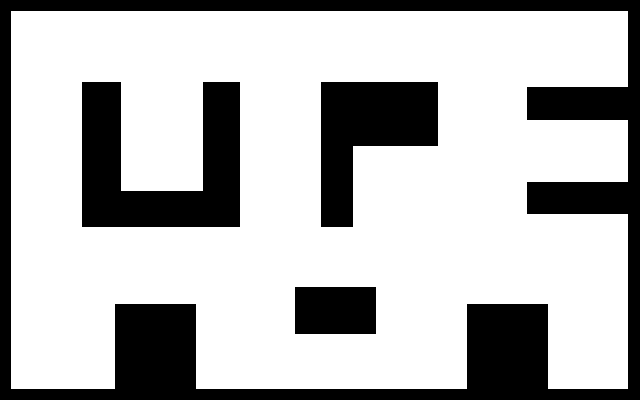}\hfil
\includegraphics[width=0.24\linewidth]{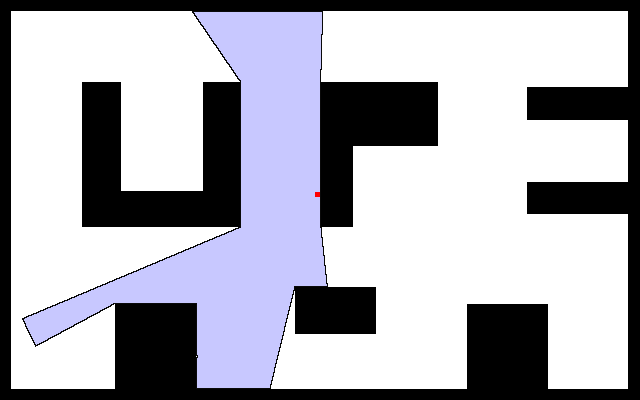}\hfil
\includegraphics[width=0.24\linewidth]{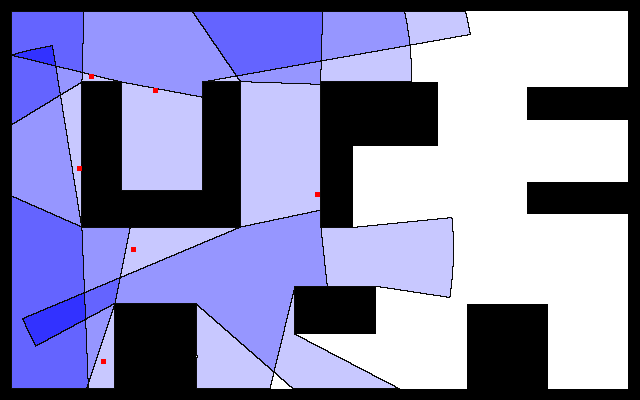}\hfil
\includegraphics[width=0.24\linewidth]{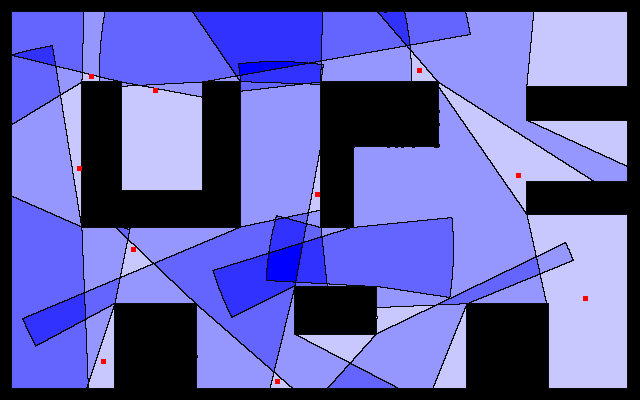}\hfil
\end{minipage}
}\par

\subfloat[]{
\begin{minipage}{\linewidth}
\includegraphics[width=0.24\linewidth]{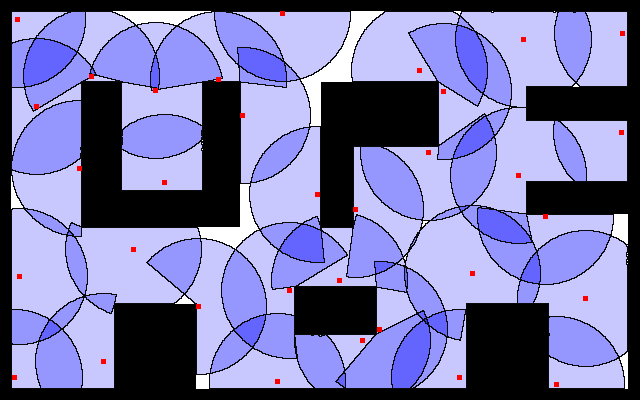}\hfil
\includegraphics[width=0.24\linewidth]{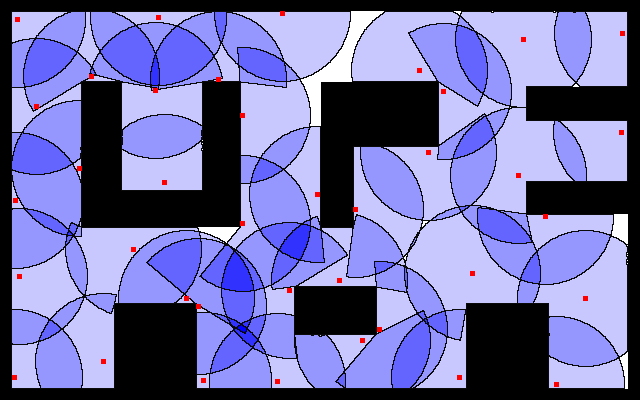}\hfil
\includegraphics[width=0.24\linewidth]{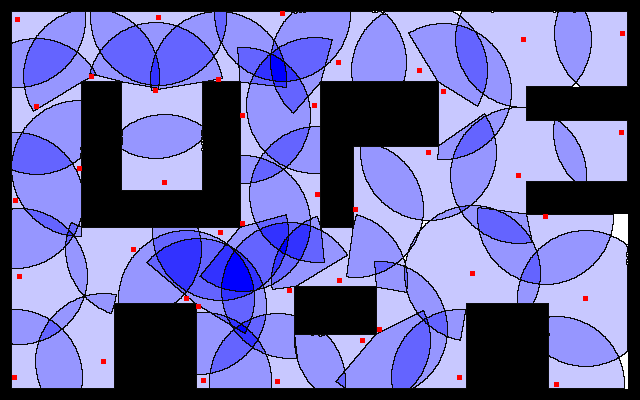}\hfil
\includegraphics[width=0.24\linewidth]{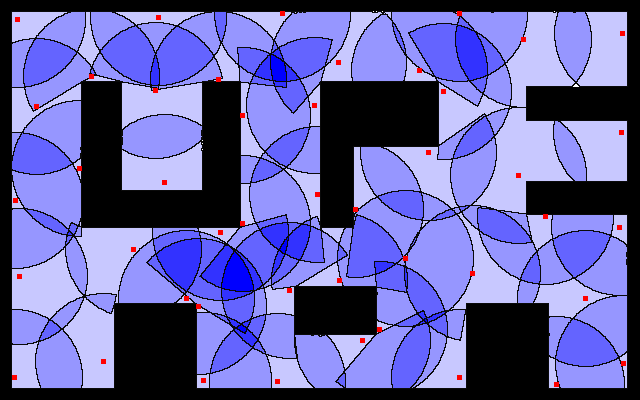}\hfil
\end{minipage}
}\par

\subfloat[]{
\begin{minipage}{\linewidth}
\includegraphics[width=0.24\linewidth]{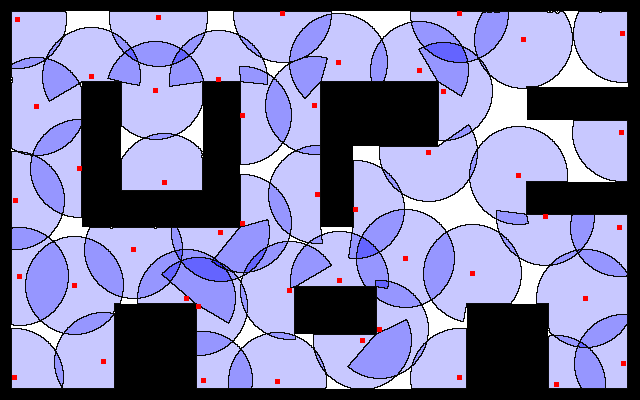}\hfil
\includegraphics[width=0.24\linewidth]{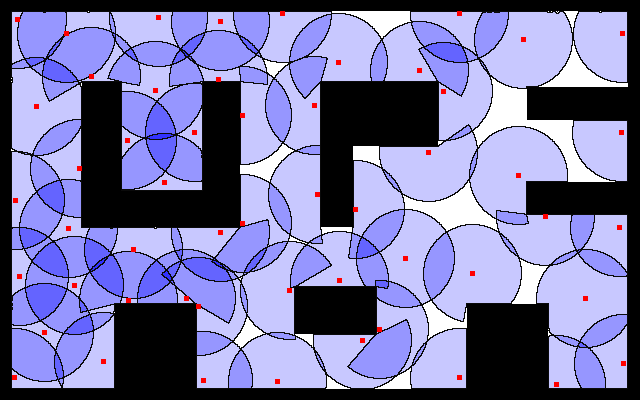}\hfil
\includegraphics[width=0.24\linewidth]{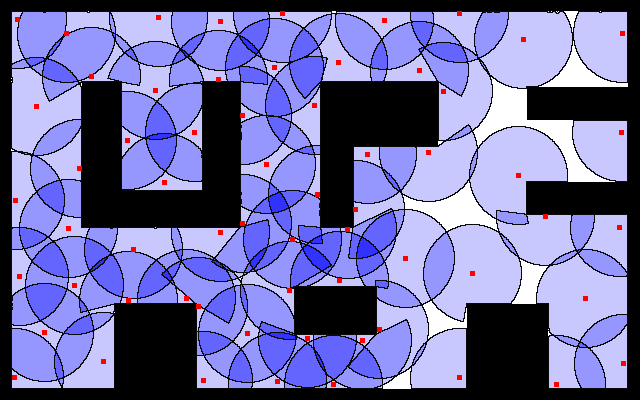}\hfil
\includegraphics[width=0.24\linewidth]{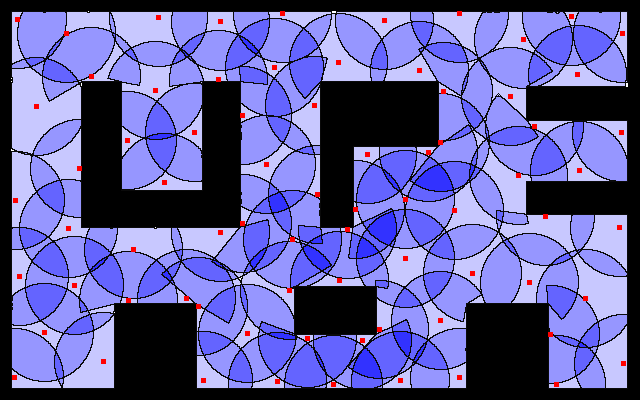}\hfil
\end{minipage}
}\par

\subfloat[]{
\begin{minipage}{\linewidth}
\includegraphics[width=0.24\linewidth]{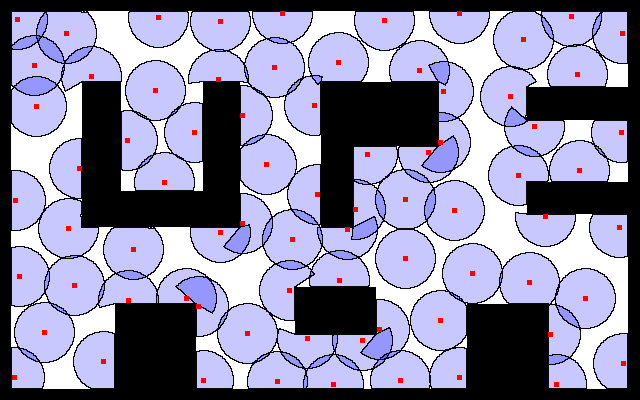}\hfil
\includegraphics[width=0.24\linewidth]{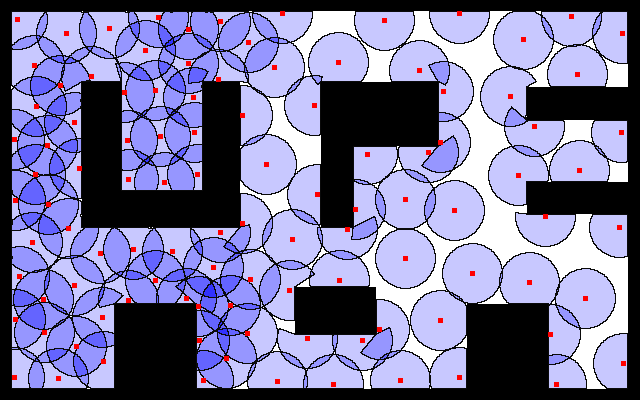}\hfil
\includegraphics[width=0.24\linewidth]{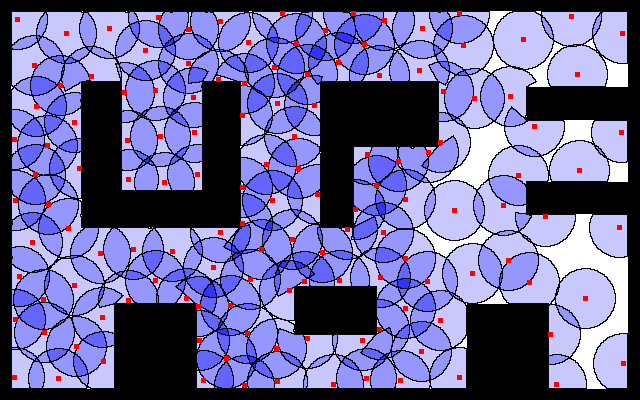}\hfil
\includegraphics[width=0.24\linewidth]{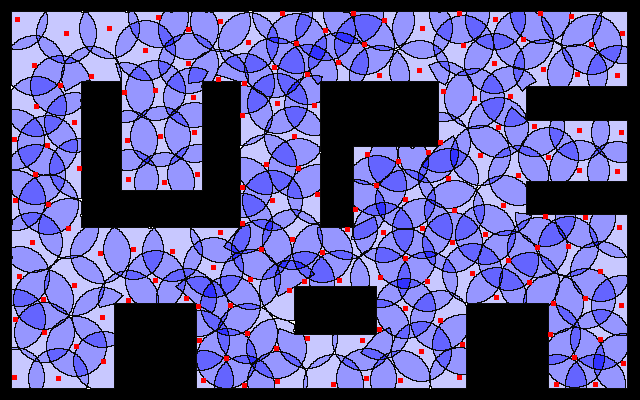}\hfil
\end{minipage}
}\par
\caption{An example of $LPA$ in ${\mathbb{R}^2}$ with omni-directional sensor footprints. Each row demonstrates populating the environment with landmarks for a fixed sensor
footprint (from a chosen filtration over the sensor footprints) at a particular iteration. In the first row sensor footprint is $S^1$ and in the last row it is the target footprint $S^s$.}
\label{Landmark Placement Algorithm}
\end{figure*}

\begin{figure}[t]
\centering
\includegraphics[width=2in]{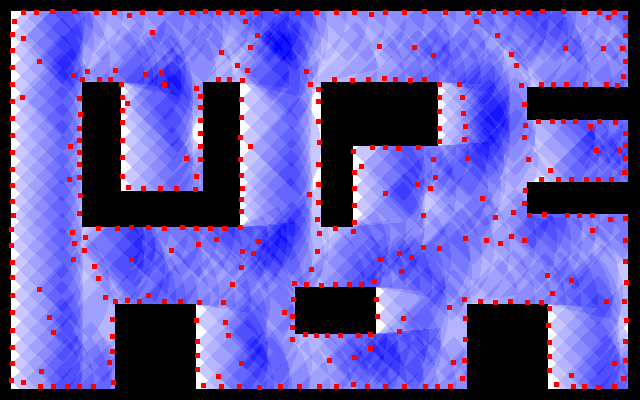}
\caption{Landmark placement algorithm final result for a simple environment in $SE(2)$.}
\label{fig6}
\end{figure}

\begin{algorithm}
    \begin{algorithmic}[1]
      \caption{Landmark Placement Algorithm} \label{LPA}
      \Require
      \Statex \textbf{a.} Workspace $\mathcal{W}$ = ${\mathbb{R}^2}$ - $O$
      \Statex \textbf{b.} Configuration Space $\mathcal{C}$
      \Statex \textbf{c.} Sensor Footprint $S$ = $\{S^1, S^2, S^3, ..., S^s\}$
      \Ensure
      \Statex \textbf{a.} Set of all Landmarks' Position $\mathcal{P}$
      \Statex \hrulefill
      \State Initiate Set $\mathcal{P}^0 = \{\}$
      \For {$t$ = $1, 2, 3, ..., s$}
	\State $\mathcal{D}_{\mathcal{P}^t, S^t}$ = $\bigcup\limits_{p \in \mathcal{P}^t} \mathcal{D}_{p, S^t}$
	\Statex
	\State $\mathcal{U}^t$ = $\mathcal{C}$ - $\mathcal{D}_{\mathcal{P}^t, S^t}$
	\While {$\mathcal{U}^t \neq \emptyset$}
	  \State $\bar{U}^t$ = $detect \_ connected \_ components(\mathcal{U}^t)$
	  \State $p$ = $Place \_ Landmark(S^t,~U_1^t)$
	  \State $\mathcal{P}^t \leftarrow \mathcal{P}^t \cup p$
	  \State $\mathcal{D}_{\mathcal{P}^t, S^t}$ = $\bigcup\limits_{p \in \mathcal{P}^t} \mathcal{D}_{p, S^t}$
	  \Statex
	  \State $\mathcal{U}^t$ = $\mathcal{C}$ - $\mathcal{D}_{\mathcal{P}^t, S^t}$
	\EndWhile
      \EndFor
      \State \Return $\mathcal{P}^s$
    \end{algorithmic}
\end{algorithm}

\begin{figure}[t]
\centering
\subfloat[First complex environment]{\includegraphics[width=1.30in]{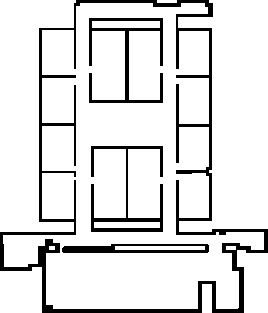}
\label{L457_orig}}\hfil
\subfloat[Landmarks populated in the first environment]{\includegraphics[width=1.30in]{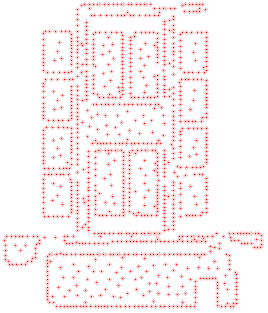}
\label{L457_LPA}}\hfil
\subfloat[Second complex environment]{\includegraphics[width=1.10in]{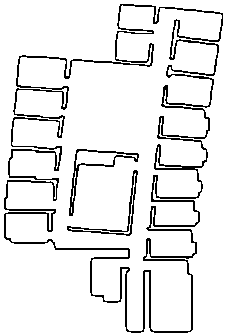}
\label{BC_orig}}\hfil
\subfloat[Landmarks populated in the second environment]{\includegraphics[width=1.10in]{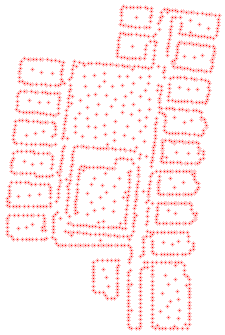}
\label{BC_LPA}}\hfil \newline
\subfloat[\u{C}ech Complex constructed by the visibility domain of the landmarks for the first environment]{\includegraphics[width=1.30in]{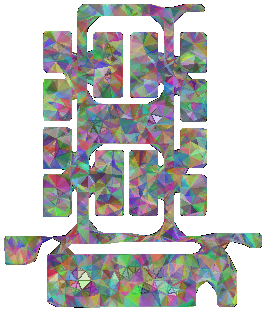}
\label{L457_cech}}\hfil
\subfloat[\u{C}ech Complex constructed by the visibility domain of the landmarks for the second environment]{\includegraphics[width=1.10in]{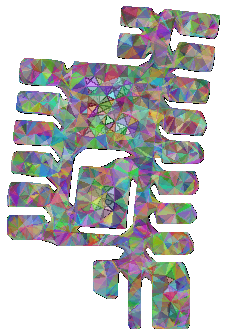}
\label{BC_cech}}\hfil
\caption{Landmark Placement Algorithm results}
\label{LPA_results}
\end{figure}

\changed{
In Algorithm \ref{LPA} the Landmark Placement Algorithm ($LPA$) is presented. $LPA$ takes the sequence of sensor footprints $S$ as an input, along with the workspace $\mathcal{W}$ and configuration space
$\mathcal{C}$. At $t^{th}$ iteration, $LPA$ computes the uncovered area $\mathcal{U}^t$ (Lines 3 and 4). While the uncovered area $\mathcal{U}^t$ is not fully covered with the union of all landmarks
visibility domains $\mathcal{D}_{\mathcal{P}^t, S^t}$, $LPA$ detects the connected components $\bar{U}^t$ and places a landmark at $U_1^t$ (Lines 6 and 7) and inserts the placed landmark into the set of
all landmarks at $t^{th}$ iteration $\mathcal{P}^t$ (Line 8). Afterwards, the algorithm updates $\mathcal{U}^t$ (Lines 9 and 10) and repeats this process until $\mathcal{U}^t = \emptyset$.

%It should be noted that, we used the pixel based representation of the image processing
%package $OpenCV$ to detect connected components of the uncovered area.

%For the case $\mathcal{C} = \mathbb{R}^2$, where the sensor footprints sequence are in shape of a disk, the direction of these disk-shaped sensor footprints does not matter. However, when
%$\mathcal{C} = SE(2)$ and the sensor footprints have a sector of a disk shape, the directionality should be taken into account. We considered this issue by breaking the domain of visibility into
%$\theta$-slices.

For the case $\mathcal{C} \subset \mathbb{R}^2$, we used the pixel based representation of the image processing package \emph{Open-CV} to detect connected components of the uncovered area. Moreover, the
$Place \_ Landmark(S^t, U_1^t)$ function, places a landmark at the centroid of the $U_1^t$ such that the visibility domain of the placed landmark $\mathcal{D}_{p^t, S^t}$ covers $U_1^t$ as much as
possible.

On the other hand, when $\mathcal{C} \subset SE(2)$, $LPA$ detects a pixel based representation of $\theta$-slice for the uncovered region $\mathcal{U}^t$ and for each $\theta$-slice, it detects the
connected components $\bar{U}^t$. In other words, the lines 6 through 10 in Algorithm \ref{LPA} will be repeated for each $\theta$-slice when $\mathcal{C} \subset SE(2)$. Moreover, the algorithm finds the
centroid of the $\theta$-slice and the $Place\_ Landmark(S^t, U_1^t)$ function places a landmark on a fixed distance $\Delta S$, away from the computed centroid in the opposite direction of $\theta$. The
reason of doing this, is to enable the visibility domain of the placed landmark $\mathcal{D}_{p^t, S^t}$ to cover the $U_1^t$ to the maximum. In order to find the optimum value of $\Delta S$ for minimum
number of landmarks placed in an environment, we tested $LPA$ with different values of $\Delta S$ which the results are presented in Figure \ref{LPA_Minimize}.

Figure \ref{Landmark Placement Algorithm}, shows an example over different iterations of $LPA$ to populate a simple environment with landmarks for $\mathcal{C} \subset \mathbb{R}^2$. In this figure, images
on each row represent one iteration on filtration over sensor footprint. Moreover, Figure \ref{fig6} shows the same environment populated with landmarks using $LPA$ where $\mathcal{C} \subset SE(2)$.
In Figure \ref{LPA_results}, results of performing the Landmark Placement Algorithm on two different complex environments are shown where $\mathcal{C} \subset SE(2)$. Figures \ref{L457_LPA} and
\ref{BC_LPA} show the placed landmarks in each environment. Figures \ref{L457_cech} and \ref{BC_cech} depict the visual representation of \u{C}ech complex constructed by considering every
$(x, y, \theta)$ in the configurations space.

}

%\begin{table}[h!]
%\begin{center}
%\caption{Number of landmarks for different $\Delta S$ values.}
%\label{fig4}
%\begin{tabular}{|c||c|} % <-- Alignments: 1st column left, 2nd middle and 3rd right, with vertical lines in between
%\hline
%\textbf{$\Delta S$ Value} & \textbf{Number of Landmarks}\\
%\hline
%0 & 2458\\
%2 & 1695\\
%4 & 1259\\
%6 & 1177\\
%8 & 1136\\
%10 & 1167\\
%12 & 1131\\
%14 & 1153\\
%18 & 1218\\
%\hline
%\end{tabular}
%\end{center}
%\end{table}

\section{Multi Robot Exploration and Navigation} \label{sec:exploration-navigation}

In this section we describe algorithms for constructing the Landmark complex through exploration of the environment, as well as algorithms for exploiting the constructed Landmark complex for informed exploration and navigation within the environment. %during the robot exploration using the landmarks scattered throughout the environment.

\subsection{Sensor Model}
At any instance of time, a robot can detect a landmark, if it falls into the robot's sensor footprint. However, the robot cannot measure the bearing or the distance to the landmarks. The only information a
robot has, is whether the detected landmarks are to its left or to its right side. This is a model for two low cost sensors attached to each side of the robot, with each sensor being able to detect the
presence of a landmark only when it is present on the sensor's side of the robot.
No other range or bearing information is assumed to be available.

\subsection{Landmark Observation}
Assuming $\mathcal{C} \subset SE(2)$ (for directional sensors) and $\mathcal{W} \subset \mathbb{R}^2$, at time $t$ the location of $i^{th}$ robot is denoted by $r_i^t = (x_i, y_i, \theta_i)$ and
$\mathcal{R}^t = \{ r_i^t \in \mathcal{C} \mid 1 \leq i \leq N \}$ is the set of all robots' locations. At each time step, each robot will make an observation to create a simplex from the observed
landmarks. Every time a robot observes a $n$-tuple of landmarks, it inserts the corresponding $(n$-$1)$-simplex constituting of the observed landmarks in $C_{n-1}$ (along with all its faces and sub-faces
in $C_i, i<n-1$).
Algorithm \ref{make_obs} takes the robot's location $r_i = (x_i, y_i, \theta_i)$ as an input. Each time the $i^{th}$ robot will detect landmarks falling into its sensor footprint and store them in
the simplex set $\mathcal{S}$. The subroutine $Make\_Observation(r_i)$ inserts not only the observed simplex $\mathcal{S}$, but also all of the faces and sub faces of $\mathcal{S}$ into the landmark
complex $\mathcal{K}$ recursively. Moreover, the landmark complex $\mathcal{K}$ is a global variable stored on a centralized server.
\begin{algorithm}
	\begin{algorithmic}[1]
		\caption{$Make\_Observation(r_i)$ \textcolor{gray}{~~//Updates the Landmark Complex, $\mathcal{K}$, using observations.}} \label{make_obs}
		\Require
		\Statex \textbf{a.} Robot's location $r_i = (x_i, y_i, \theta_i)$
		\Statex \textbf{b.} Landmark Complex $\mathcal{K}$ (Global Variable)
		\Statex \hrulefill
		\State Set $\mathcal{S} = Detect\_Landmarks(r_i)$ %\Statex
		\hfill\textcolor{gray}{//Set of $0$-simplices observed by $i^{th}$ robot at $(x_i, y_i, \theta_i)$}
		\State $\mathcal{K} = Update\_Landmark\_Complex(\mathcal{S})$ %\Statex
		\hfill\textcolor{gray}{//Recursively insert simplex $\mathcal{S}$ and its faces to $\mathcal{K}$} 
	\end{algorithmic}
\end{algorithm}

\subsection{Robot Short-Term-Trajectory ($STT$) Modeling}

In this section, a non-holonomic model for generating robot short-term-trajectories is presented, which can be used for directing the robot towards a specific landmark in its domain of visibility.
This short-term-trajectory generation will be used as a low-level controller for for the different \emph{modes of walk} described in Section~\ref{sec:modes-of-walk}.

\begin{figure}[t]
\centering
\includegraphics[width=3in]{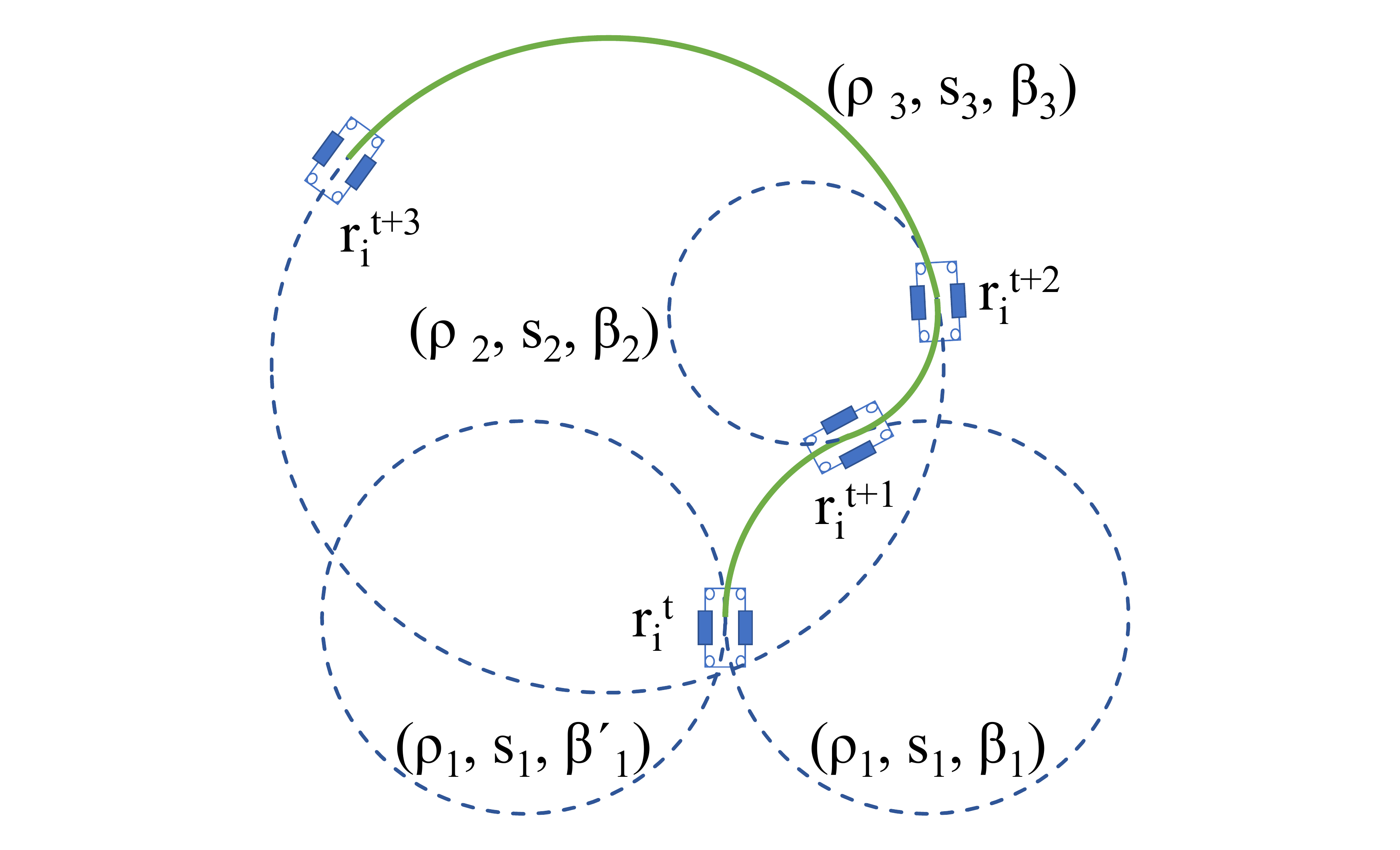}
\caption{\textbf{Robot Short-Term Trajectory Modeling: }This example, shows a robot following the trajectories modeled as Dubins curves to reach a goal point.}
\label{DubinsCurves}
\end{figure}

Since the location of landmarks and robots are unknown in the environment, our strategy is to generate
short paths (short-term-trajectories) described by Dubins curves\cite{Dubins1957OnCO} for each robot in order to attain a short term control objective. In other words, we modeled the short-term-trajectories of
the robots as arcs of circles with distinct radius $\rho$ and arc length $s$, tangent to the robot's current orientation (See Figure \ref{DubinsCurves}). Given these two parameters, there would be two
choices of Dubins curves. One make the robot to turn towards its left and the other one to its right. This binary choice is described by a two-state variable $\beta$ which can either be +1 (right turn) or
-1 (left turn). Consequently, the short-term-trajectories are described by three variables, $(\rho, s, \beta)$.

It is notable that $\rho$ and $s$ are limited to upper bounds and lower bounds. $\rho$ can assume value in range of $[0,~\infty)$ where $\rho = 0$ allowing rotations at the
robot's position and $\rho = \infty$ corresponding to a straight line. However for computational purposes, we assume a large finite $\rho_{max}$ as the upper bound on $\rho$. Moreover, $s$ can assume value
in range of $[0,~\min\{s_{max},~\pi \rho\})$ where $s_{max}$ is a constant value. Choosing the upper bound from the minimum of $s_{max}$ and $\pi \rho$ will avoid arc lengths grater than half a circle.
Hence, $\rho$ will have a value from $[0,~\rho_{max})$ and $s$ from $[0,~\min\{s_{max},~\pi \rho\})$.

We assume robots can detect obstacles when they are in contact with them. In such case, if robots collide with an obstacle while executing the short-term-trajectory, they will turn on their position to
acquire a new orientation and resume exploring.
Figure \ref{DubinsCurves} shows an example of robot short-term-trajectory sequence modeled as Dubins curves. In this example, the position of the $i^{th}$ robot at the $t^{th}$ time step is denoted as
$r^{t}_{i}$. At each time step, robot will choose a Dubins curve as the short-term-trajectory and follows the created path to reach the goal. In this case the followed path is
$\mathcal{T} = \{(\rho_1, s_1, \beta_1), (\rho_2, s_2, \beta_2), (\rho_3, s_3, \beta_3)\}$. In this figure the path $(\rho_1, s_1, \beta^{'}_{1})$ also can be found which implies that with a same
$\rho_1$ and $s_1$ there exists two paths. This shows the importance of $\beta$.

\subsection{Modes of Walk} \label{sec:modes-of-walk}

In this section we describe different modes of walk that the robots can assume depending on the desired balance between exploration of the unknown regions of the environment and exploitation of the partially-constructed Landmark complex.
These include \emph{random walk} which is the only feasible option when the the robots know nothing about the environment or when the robots are venturing into completely unexplored sections of the environment.
But with a partially-constructed Landmark complex the robots can exploit it to perform an \emph{Informed Systematic Walk} for more efficient exploration of the environment at the frontiers to the unexplored regions.
Finally, when the Landmark complex is mostly complete, we can use tools from homology theory to detect ``\emph{holes}'' in complex (small islands of unexplored regions) and use the robots to perform targeted exploration of such regions in order to complete the Landmark complex using a \emph{Homology Informed Walk}.

\subsubsection{{Random Walk}}
Since at the beginning of the exploration the Landmark complex is empty and there is no reference for the robots to use as a guide (recall that the location of landmarks and robots are unknown), robots need to perform \emph{Random Walk} ($RW$) to construct a partial Landmark complex in order to localize
themselves with respect to the observed landmarks. In this section, a description of the Random Walk algorithm is provided.

\begin{figure}[t]
\centering
\includegraphics[width=3in]{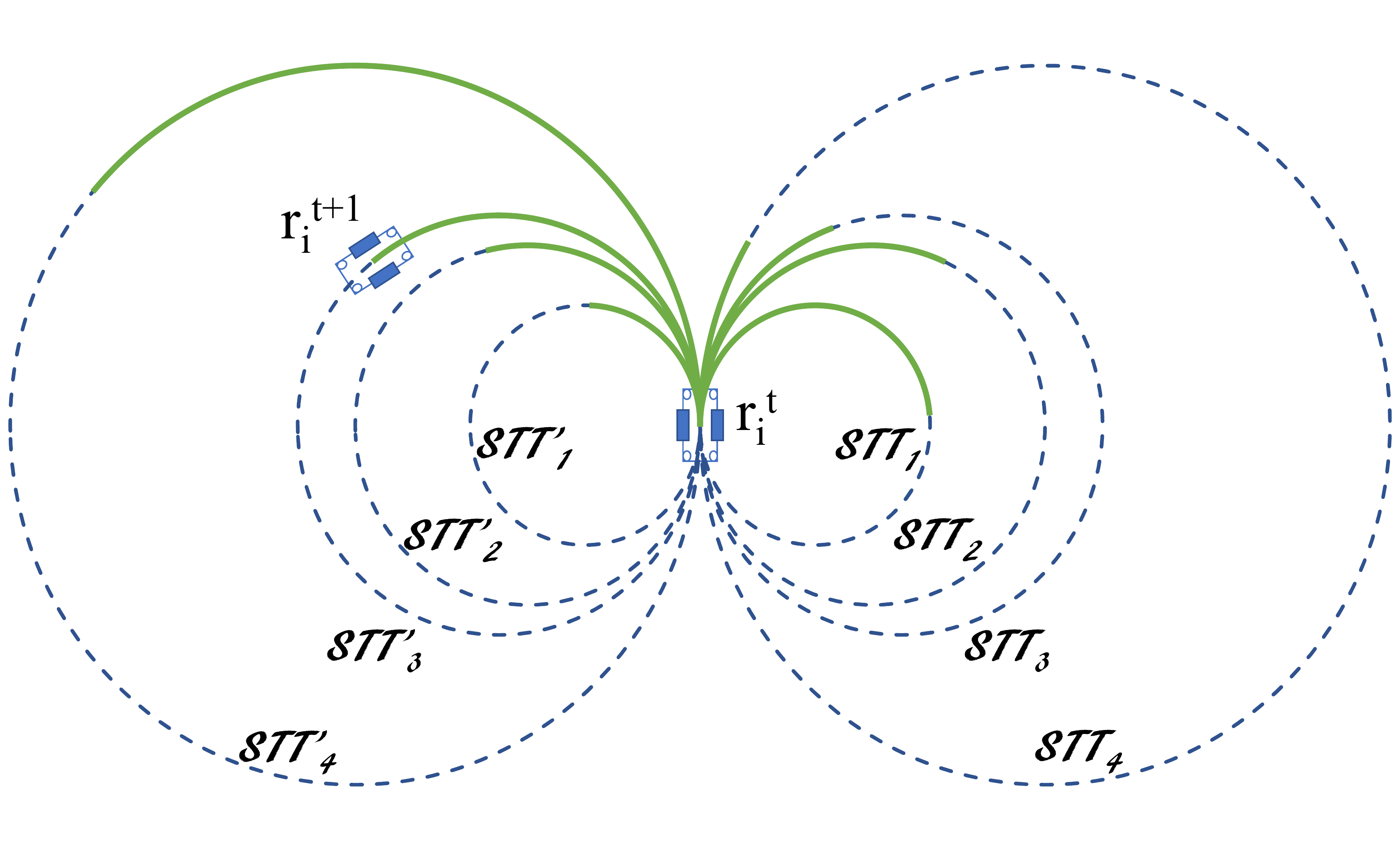}
\caption{An illustrative example of Random Walk algorithm. In this figure, different short-term-trajectories ($STT$) have been shown where $STT_i$ is $(\rho_i, s_i, \beta_i)$ and $STT^{'}_i$ is
$(\rho_i, s^{'}_{i}, \beta^{'}_{i})$.}
\label{figRW}
\end{figure}

To develop a Random Walk, all three variable for generating the robots' short-term-trajectory $(\rho, s, \beta)$, are randomly chosen with $\rho$ and $s$ being selected from a uniform probability
distribution. Therefore, $\rho$ will be sampled from $[0,~\rho_{max})$ and $s$ from $[0,~\min\{s_{max},~\pi \rho\})$. This will enable the robots to choose between a vast variety of paths at each time
step. For instance Figure \ref{figRW} shows some of these possible paths. In this figure, in time $t$, the $i^{th}$ robot is in $r^{t}_{i}$ and randomly picks one path which in here is
$(\rho_3, s^{'}_{3}, \beta^{'}_{3})$ and moves to $r^{t+1}_{i}$.
We refer to a single step of Random Walk as the execution of single short-term-trajectories explained above.

\begin{algorithm}
  \begin{algorithmic}[1]
  \caption{$r_i^* := RW\_Observe(r_i)$} \label{RW_algo}
  \Require
  \Statex \textbf{a.} Robot's location $r_i = (x_i, y_i, \theta_i)$
  \Ensure
  \Statex \textbf{a.} Robot's updated location $r_i^* = (x_i^*, y_i^*, \theta_i^*)$
  \Statex \hrulefill
  \State $[\rho,~s,~\beta] := UniRand\_Sample(\rho_{max},~s_{max})$
  \State Set $\mathcal{J} := Generate\_Path(\rho,~s,~\beta)$
  \For {$k \in (1, 2, \cdots, \lvert \mathcal{J} \rvert$)}
    \State $r_i \leftarrow \mathcal{J}[k]$ \null\hfill\textcolor{gray}{//$\mathcal{J}[k] = (x_i^k, y_i^k, \theta_i^k) \in \mathcal{C}$}
    \State $Make\_Observation(r_i)$
  \EndFor
  \State \Return $r_i^*$
    \end{algorithmic}
\end{algorithm}

\changed{
The pseudo code given in Algorithm \ref{RW_algo}, describes the Random Walk and Observation subroutine $RW\_Observe(r_i)$. The input to this function is the $i^{th}$ robot's location $r_i$. It will move
the robot to the new random location $r_i^*$ while observing new simplices and updating the landmark complex $\mathcal{K}$. In line 1, the variables ($\rho,~s,~\beta$) are sampled randomly from a uniform
probability distribution. Furthermore, the function $Generate\_Path$ returns a set of points $\mathcal{J}$ in the configuration space $\mathcal{C} \subset SE(2)$ (Line 2). The $i^{th}$ robot's location
is updated with these configurations, while the robot makes an observation to update $\mathcal{K}$ each time along the path (Lines 3 to 6).
}

\subsubsection{{Informed Systematic Walk}}
Once robots have created a partial Landmark complex, they can exploit that information to perform a more systematic form of walk which result in faster landmark complex construction through increased exploration at the boundaries/frontiers of the unexplored regions.

The key insight behind this mode of walk is to navigate the robots to the landmarks that are observed fewer times in comparison to other landmarks. Those landmarks correspond to regions that are more
likely to have remained unexplored (frontier or boundary regions). We refer to these landmarks as \emph{frontier landmarks}. In other words, frontier landmarks are expected to be the boundaries of the unexplored regions since they are
observed fewer number of times. The Informed Systematic Walk $(ISW)$ is designed to navigate the robots to the least observed landmarks. Moreover, $ISW$ partitions a set of landmarks based on the number of
robots and navigates each robot to the least observed landmark within its partition. In following subsections, the descriptions of individual components of $ISW$ are given.

\textbf{\emph{i.} Voronoi partitioning based landmark assignment}:
Since there are multiple robots, we used graph search-based Voronoi partitioning to assign landmarks to the robots. In other words, $ISW$ will partition the environment into cells centered around the
robots and assign the least observed landmark in each cell to the corresponding robot.

We construct the graph $G$ from the partial Landmark complex built by robots during the exploration, such that the vertex set $\mathcal{V}(G)$ is the $0$-simplices and the edge set $\mathcal{E}(G)$ is the
$1$-simplices of the constructed landmark complex. We use a Dijkstra type wave front propagation algorithm to construct Voronoi Partitioning on the graph $G$~\cite{coverage:riemannian:IJRR:13}. In order to do so, we initiate the
\emph{open list} with the vertices corresponding to the landmarks currently being observed by the robots. Moreover, the cost on every edge $(C_G)$ is equal to 1 unit.
The pseudo-code given in Algorithm \ref{VP_algo}, describes our Voronoi partitioning algorithm which returns the tessellation map, $\uptau$, of landmarks to the robots. In line 6 the \emph{open list} is
initiated with the landmarks observed by the $i^{th}$ robot at current time step for every $i = 1, 2, \cdots, N$ and in line 7 these landmarks are assigned a partition identity of $i$ which means the
landmark is assigned to the $i^{th}$ robot. Lines 11 through 21 corresponds to the main loop of the algorithm. With every iteration, the vertex with minimum $g$-score in $Q$ (the unexpanded vertices), is
expanded. Furthermore, the algorithm checks for improvement in $g$-score for the neighboring vertices of the expanded vertex (line 14 to 20). If the potential $g$-score of a neighbor is less than the
current one, the algorithm updates the $g$-score of that vertex and when a vertex is updated with an improved $g$-score, the tessellation identity of that vertex also gets updated (line 18).

\begin{figure}
\centering
\includegraphics[width=2.75in]{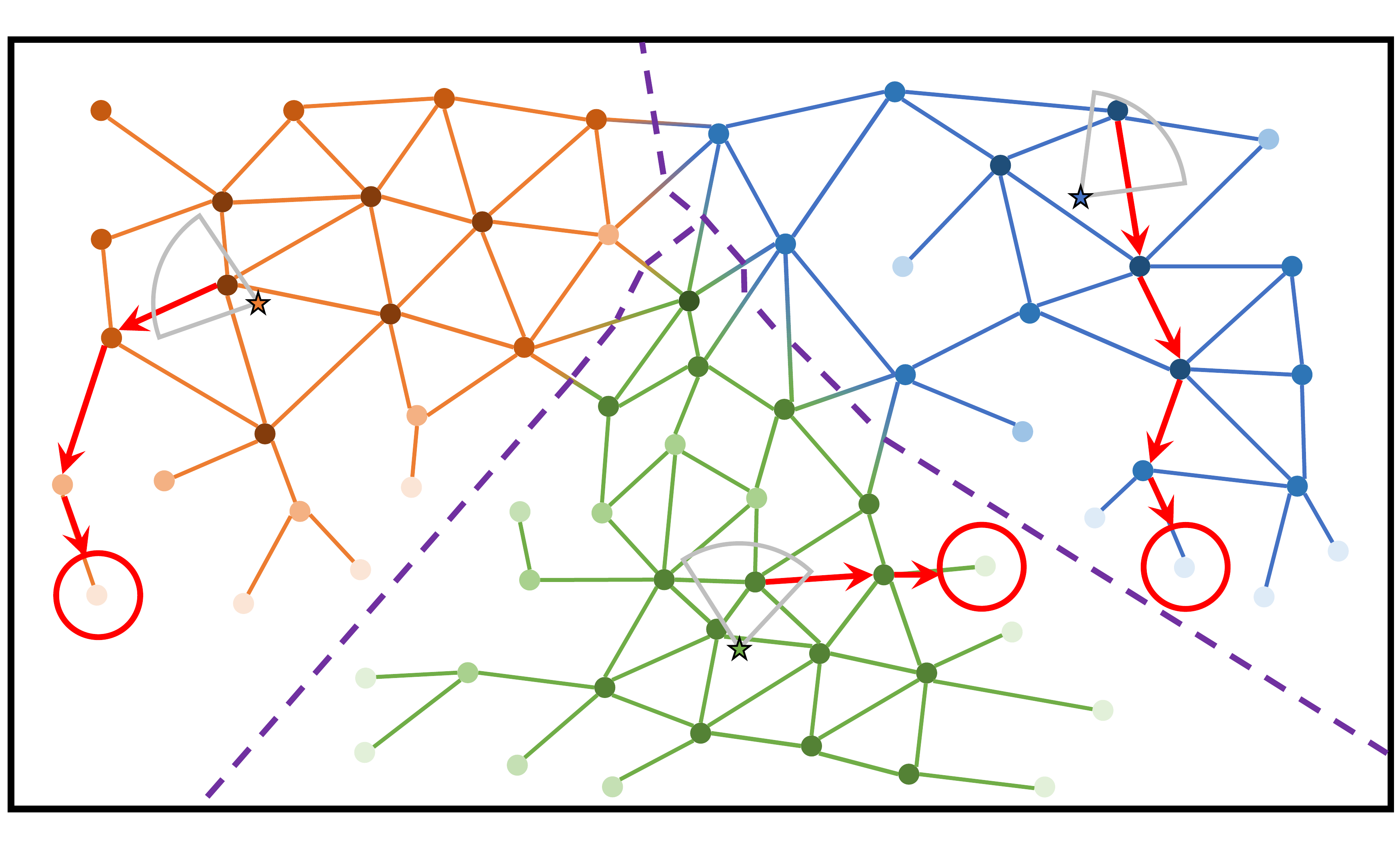}
\caption{Voronoi Partitioning around each robot (depicted as stars). Landmarks with darker colors imply that they have been observed more than landmarks with
lighter colors. Landmarks that are observed fewer number of times form the boundary are the $ISW$ target.}
\label{figVP}
\end{figure}

\begin{algorithm}
    \begin{algorithmic}[1]
      \caption{Voronoi Partitioning Algorithm \newline $\uptau := Voronoi(\mathcal{R},~G)$} \label{VP_algo}
      \Require
      \Statex \textbf{a.} Graph $G$ (with vertex set $\mathcal{V}(G)$, edge set $\mathcal{E}(G) \subseteq \mathcal{V}(G) \times V(G)$ and cost function $C_G$)
      \Statex \textbf{b.} Set of Agents location: \Statex $\mathcal{R} = \{r_i \in \mathcal{C} \subset SE(2) \mid 1 \leq i \leq N\}$
      \Ensure
      \Statex \textbf{a.} The tessellation map $\uptau:\mathcal{V}(G) \rightarrow \{1, 2, \cdots, N\}$
      \Statex \hrulefill
      \State Initiate $g$: Set $g(v) := \infty, \forall v \in \mathcal{V}(G)$ \null\hfill\textcolor{gray}{//Shortest distances}
      \State Initiate $\uptau$: Set $\uptau (v) := -1 , \forall v \in \mathcal{V}(G)$ \null\hfill\textcolor{gray}{//Tessellation}
      \State Detect landmarks for $i^{th}$ robot, $\{p_{i, 1}, \cdots, p_{i,M_i}\},~i = 1, 2, \cdots, N$.
      \null\hfill\textcolor{gray}{//$N$ is the number of robots and $M_i$ is the number of landmarks observed by each robot}
      \For {$i \in \{1, \cdots, N\}$}
	\For {$k \in \{1, \cdots, M_i\}$}
	  \State Set $g(p_{i, k}) := 0$
	  \State Set $\uptau(p_{i,k}) := i$
	\EndFor
      \EndFor
      \State Set $Q := \mathcal{V}(G)$ \null\hfill\textcolor{gray}{//Set of unexpanded vertices}
      \While {$Q \neq \emptyset$}
	\State $q := $argmin$_{q^{\prime} \in Q}~g(q^{\prime})$ \Statex\hfill\textcolor{gray}{//Maintained by a heap data structure}
	\State Set $Q = Q - q$  \null\hfill\textcolor{gray}{//Remove $q$ from $Q$}
	\For {\textbf{each} $\{w \in \mathcal{N}_G(q)\}$} \null\hfill\textcolor{gray}{//For each neighbor of $q$}
	  \State Set $g^{\prime} := g(q) + C_G([q,w])$
	  \If {$g^{\prime} < g(w)$}
	    \State Set $g(w) := g^\prime$
	    \State Set $\uptau(w) := \uptau(q)$
	  \EndIf
	\EndFor
      \EndWhile
      \State \Return $\uptau$
    \end{algorithmic}
\end{algorithm}

Figure \ref{figVP} depicts Voronoi partitioning and shows the path that the robots need to take in order to reach the least observed landmark in the corresponding partition.

\textbf{\emph{ii.} Dijkstra's search for shortest path in 1-skeleton}:
Once the tessellation $\uptau$ is computed, each robot find the landmark with least observation count within its own partition. We refer to this landmark as \emph{goal landmark}. We construct the shortest
path from the robots' current location to the \emph{goal landmark} using Dijkstra algorithm on the graph $G$. The output of Dijkstra search algorithm is a sequence of landmarks that a robot need to
navigate along. Suppose the sequence of landmarks (\emph{path landmarks}) for $i^{th}$ robot is $L_i = \{S_i, l_1, l_2, \cdots, G_i\}$. Each robot will use the obtained \emph{path landmarks} to reach its
assigned \emph{goal landmark} in its own partition.

\begin{algorithm}
    \begin{algorithmic}[1]
      \caption{Informed Systematic Walk Algorithm \newline $[i^*,~g\_score] := ISW(\mathcal{R},~i)$} \label{ISW_algo}
      \Require
      \Statex \textbf{a.} Set of Agents location: \Statex $\mathcal{R} = \{r_i \in \mathcal{C} \subset SE(2) \mid 1 \leq i \leq N\}$
      \Statex \textbf{b.} Identity of the robot $i$
      \Statex \textbf{c.} Landmark Complex $\mathcal{K}$ (Global Variable)
      \Ensure
      \Statex \textbf{a.} Identity of the goal landmark $i^*$
      \Statex \textbf{b.} Distance to the goal landmark $g\_score$
      \Statex \hrulefill
      \State Graph $G :=$ $Construct\_Graph(\mathcal{K})$
      \State The tessellation map $\uptau := Voronoi(\mathcal{R},~G)$
      \State $i^* := Least\_Observed\_Landmark(\uptau,~i)$
      \State $g\_score := Search(i^*,~G)$
      \State \Return $[i^*,~g\_score]$
    \end{algorithmic}
\end{algorithm}

\changed{
Algorithm \ref{ISW_algo} describes our $ISW(\mathcal{R},~i)$ subroutine. Inputs to this function are the set of robots location $\mathcal{R}$ and the identity of the robot running this function $i$. It
also uses the global variable $\mathcal{K}$ to construct the graph $G$ (Line 1). In line 2, the algorithm computes the tessellation map $\uptau$ by using the $Voronoi(\mathcal{R},~G)$ subroutine, described
in Algorithm \ref{VP_algo}. Afterwards, the least observed landmark (goal landmark) in the $i^{th}$ robot partition will be identified $i^*$ (Line 3), and by using Dijkstra algorithm on graph $G$, the
shortest distance ($g\_score$) of the $i^{th}$ robot to the goal landmark $i^*$ is computed (Line 4). Therefore, the outputs of $ISW$ are the identity of the $i^{th}$ robot's goal landmark $i^*$ and the
shortest distance from the robot to its goal landmark $g\_score$.
}

\textbf{\emph{iii.} Navigation}:
In order to navigate along the path landmarks $L_i$, the $i^{th}$ robot needs to generate short-term-trajectories that will take it from one landmark in the sequence into the next.
Suppose the $i^{th}$ robot observed $l_k$ in the sequence. In order to navigate to $l_{k+1}$, it first need to make sure $l_{k+1}$ is within its sensor footprint. Based on our assumption on the sensor
model, a robot cannot measure the bearings or the distance to the landmarks. However, it has the information on whether $l_{k+1}$ is to its left or to its right side. Based on this information, robot
will choose the proper $\beta$ to create the appropriate short-term-trajectory to navigate towards its goal landmark. $\beta = +1$ if $l_{k+1}$ is to the robot's right side and $\beta = -1$ if $l_{k+1}$ is
to its left side. However, we sample $\rho$ and $s$ randomly as before.

On the other hand, if $l_{k+1}$ is not visible to the $i^{th}$ robot, few steps of random walk will be taken in order to find $l_{k+1}$. In case that even after taking few steps of random walk, $l_{k+1}$
was not visible, a new Dijkstra search will be executed to generate new sequence of path landmarks to the goal.

\begin{figure}[t]
\centering
\includegraphics[width=3in]{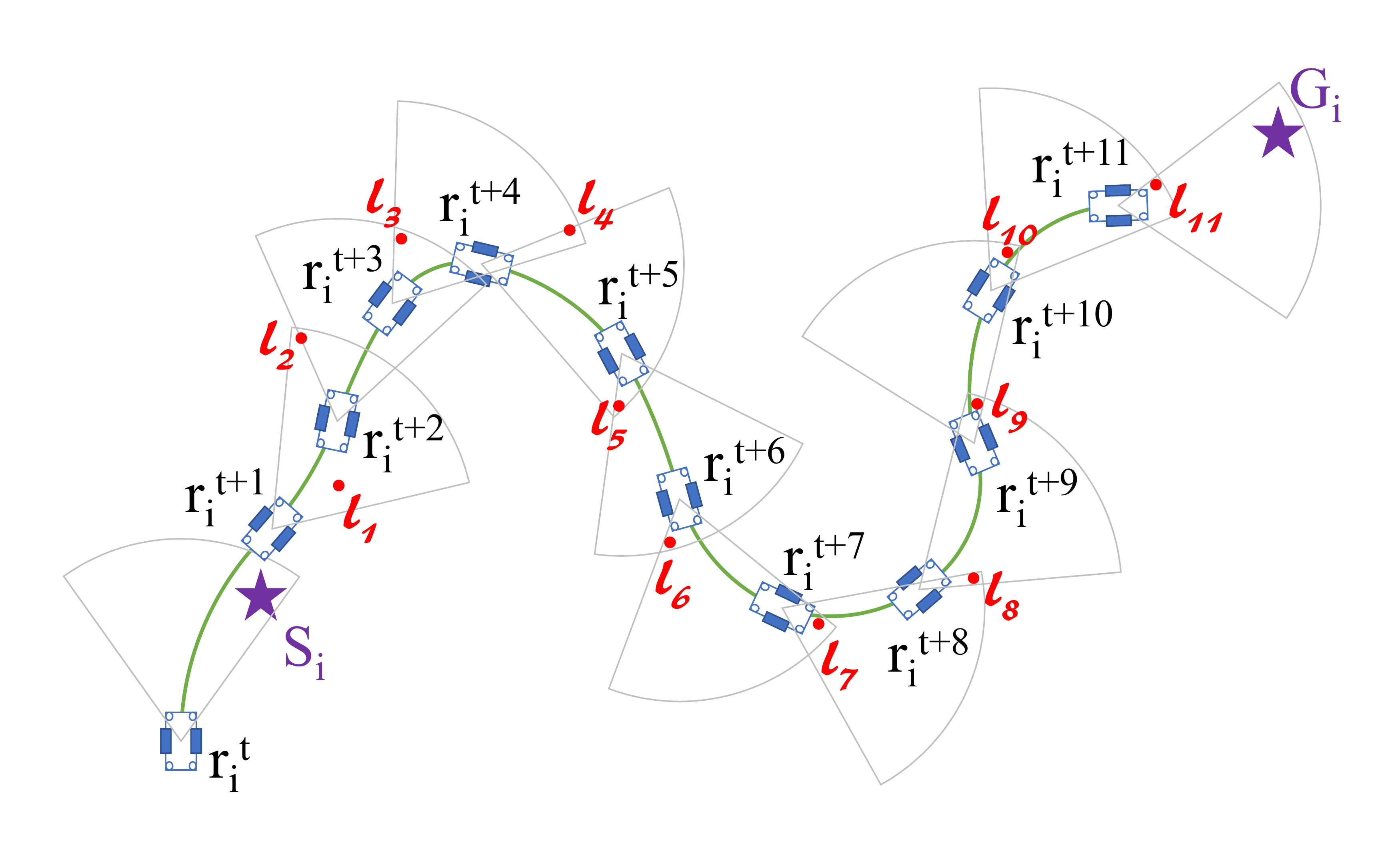}
\caption{\textbf{Navigation Algorithm: }The $i^{th}$ robot uses the path landmarks $L_i$ to navigate from $S_i$ to $G_i$.}
\label{Navigation}
\end{figure}

Figure \ref{Navigation} illustrates an example of the navigation algorithm. In this example the $i^{th}$ robot is depicted in different time steps. Moreover,
$L_i = \{S_i, l_1, l_2, l_3, l_4, ..., l_{11}, G_i\}$ is the sequence of path landmarks provided by the Dijkstra algorithm, that directs the robot from starting landmark $S_i$ towards the goal landmark
$G_i$. At each time step, robot chooses $\beta$ according to the orientation of the nearest path landmark.

\begin{algorithm}
  \begin{algorithmic}[1]
    \caption{$r_i^* := ESTT\_Observe(r_i, l_k)$} \label{STT_algo}
    \Require
    \Statex \textbf{a.} Robot's location $r_i = (x_i, y_i, \theta_i)$
    \Statex \textbf{b.} Observed landmark $l_k$
    \Ensure
    \Statex \textbf{a.} Robot's updated location $r_i^* = (x_i^*, y_i^*, \theta_i^*)$
    \Statex \hrulefill
    \State $[\rho,~s] := UniRand\_Sample(\rho_{max},~s_{max})$
    \State $\beta := Left\_Right(r_i,~l_k)$
    \State Set $\mathcal{J} := Generate\_Path(\rho,~s,~\beta)$
    \For {$k \in (1, 2, \cdots, \lvert \mathcal{J} \rvert$)}
      \State $r_i \leftarrow \mathcal{J}[k]$ \null\hfill\textcolor{gray}{//$\mathcal{J}[k] = (x_i^k, y_i^k, \theta_i^k) \in \mathcal{C}$}
      \State $Make\_Observation(r_i)$
    \EndFor
    \State \Return $r_i^*$
  \end{algorithmic}
\end{algorithm}

\changed{
In Algorithm \ref{STT_algo}, an outline of Execute Short-Term Trajectory and Observation subroutine $ESTT\_Observe(r_i,~l_k)$ is presented. Similar to $RW\_Observe(r_i)$ subroutine, this function also
takes the the $i^{th}$ robot's location $r_i$ and returns the updated location $r_i^*$ while observing new simplices and updating the landmark complex $\mathcal{K}$. As described in Algorithm
\ref{RW_algo}, $RW\_Observe(r_i)$ generates the short-term-trajectories completely randomly by sampling all $(\rho,~s,~\beta)$ variables from uniform probability distribution. However, in line 2 of
$ESTT\_Observe(r_i,~l_k)$ subroutine, the $Left\_Right$ function takes the observed landmark $l_k$ and the $i^{th}$ robot's location $r_i$ as inputs and checks whether $l_k$ is to the left or right side of
the $i^{th}$ robot and return $\beta = +1$ if $l_k$ is to the robot's right and $\beta = -1$ if it is to the left. Nevertheless, $\rho$ and $s$ are still randomly chosen (Line 1). Afterwards, similar to
$RW\_Observe(r_i)$, the set of points $\mathcal{J}$ in the configuration space $\mathcal{C} \subset SE(2)$ is generated by taking the $(\rho,~s,~\beta)$ variables (Line 3). Ultimately, for each
configuration in $\mathcal{J}$, the $i^{th}$ robot's location is updated while making observations to update $\mathcal{K}$ (Lines 4 to 7).

}

\begin{algorithm}
    \begin{algorithmic}[1]
      \caption{$r_i^* := Navigate(r_i,~i^*)$} \label{nav}
      \Require
      \Statex \textbf{a.} Robot's location $r_i = (x_i, y_i, \theta_i)$
      \Statex \textbf{b.} Identity of the goal landmark $i^*$
      \Statex \textbf{c.} Landmark Complex $\mathcal{K}$ (Global Variable)
      \Ensure
      \Statex \textbf{a.} Robot's updated location $r_i^* = (x_i^*, y_i^*, \theta_i^*)$
      \Statex \hrulefill
      \State Initiate Bool $successful$ := false
      \While {(!$successful$)}
	\State Initiate Bool $executed\_rw$ := false
	\State Graph $G :=$ $Construct\_Graph(\mathcal{K})$
	\State Set $L_i := Shortest\_Path(i^*,~G)$\null\hfill\textcolor{gray}{//}
	\While {$L_i \neq \emptyset$}
	  \State Initiate Bool $visible$ := false
	  \For {$j \in (\lvert L_i \rvert, \cdots, 1)$}\null\hfill\textcolor{gray}{//}
	    \If {$L_i[j]$ is visible to $r_i$}
	      \State $r_i \leftarrow ESTT\_Observe(r_i,~L_i[j])$
	      \For {$k \in \{1, 2, \cdots, j\}$}
		\State Set $L_i := L_i - L_i[k]$
	      \EndFor
	      \State $visible$ := true
	      \State \textbf{break}
	    \EndIf
	  \EndFor
	  \If {(!$visible$)}
	    \If {($executed\_rw$)}
	      \State \textbf{break}
	    \Else
	      \For {$q \in (1, 2, \cdots, \sigma)$}
		\State $r_i \leftarrow RW\_Observe(r_i)$
	      \EndFor
	      \State Bool $executed\_rw$ := true
	    \EndIf
	  \EndIf
      \EndWhile
      \If {$L_i = \emptyset$}
	\State Bool $successful$ := true
      \EndIf
    \EndWhile
    \State \Return $r_i^*$
    \end{algorithmic}
\end{algorithm}

\changed{
In Algorithm \ref{nav}, the pseudo-code for the $Navigate(r_i, i^*)$ subroutine is presented. Inputs of this function are the position of the $i^{th}$ robot and the identity of the goal landmark $i^*$. In
line 5, the function $Shortest\_Path$ takes the graph $G$ and goal landmark's identity $i^*$ as inputs and returns the set of path landmarks $L_i$ for the $i^{th}$ robot to the goal landmark. In lines 8
through 18 of the algorithm, $i^{th}$ robot tries to find the furthest landmark in path landmarks set $L_i$ in order to make a shortcut if there is any available (Line 8). Once the furthest visible
landmark in $L_i$ is identified ($j^{th}$ path landmark $L_i[j]$), the algorithm executes a short-term-trajectory based on the orientation of the $j^{th}$ landmark by using the $ESTT\_Observe(r_i,~L_i[j])$
subroutine described in Algorithm \ref{STT_algo} (Line 10). This function moves the $i^{th}$ robot to the next location while making observations along the path to update landmark complex $\mathcal{K}$.
Afterwards, all the path landmarks until $L_i[j]$ will be removed from $L_i$ (Lines 11 to 13). However, if non of the path landmarks in $L_i$ are visible to the $i^{th}$ robot, the algorithm will perform
few steps $(\sigma)$ of Random Walk (Lines 22 to 25) and checks for the visibility of the landmarks in $L_i$ one more time. Even if after the execution of Random Walk non of the remaining path landmarks
are visible to the robot, a new Dijkstra search will be performed to generate new set of path landmarks $L_i$.
}

\begin{figure}
\centering
\includegraphics[width=3in]{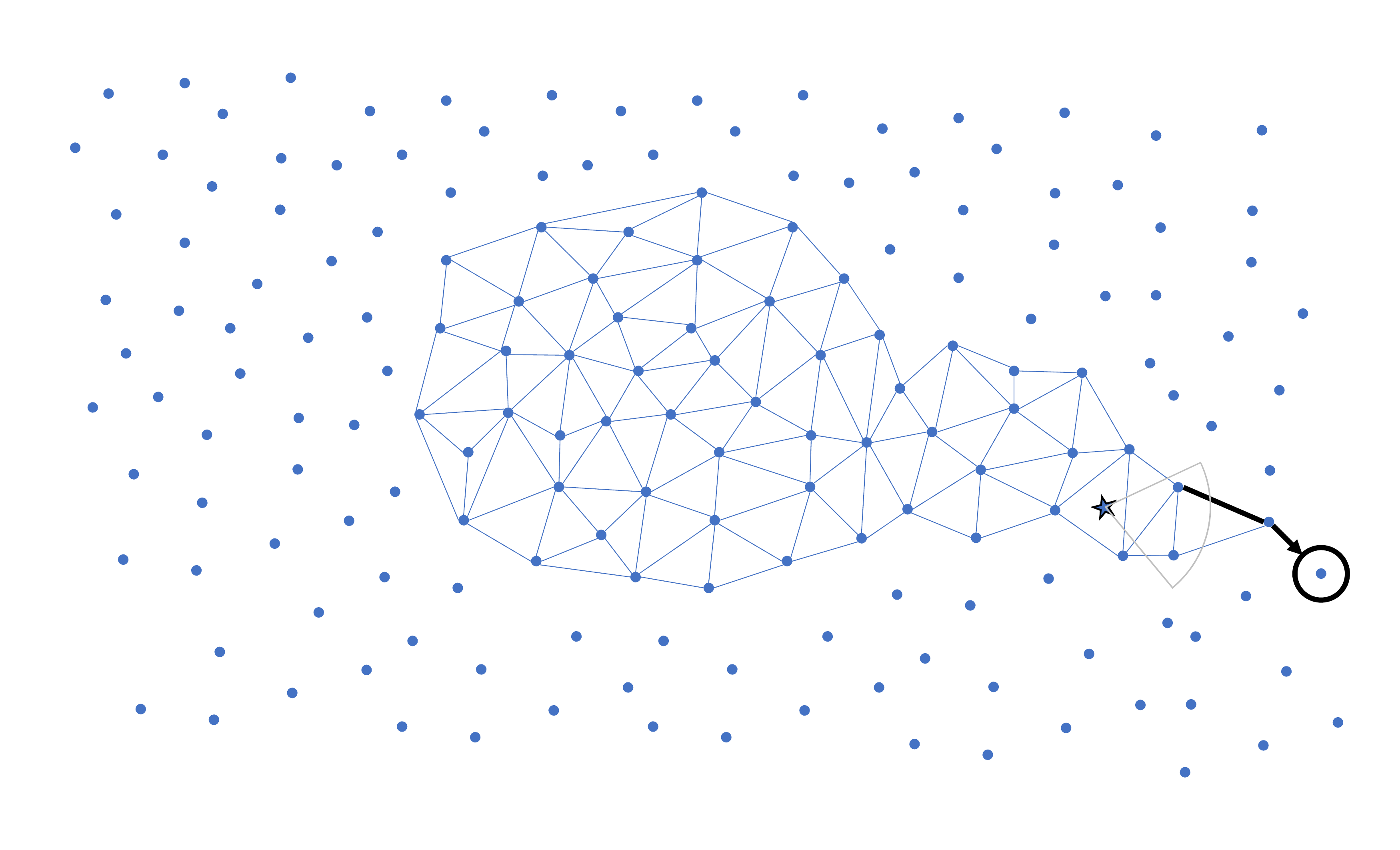}
\caption{Too many consecutive $ISW$ will lead the boundary of the simplicial complex to grow in horn-like shape as depicted.}
\label{fig12}
\end{figure}

\textbf{\emph{iv.} Interleaving between \emph{ISW} and \emph{RW}}:
A noteworthy point in $ISW$ is to not use it repeatedly. Since $ISW$ always navigates the robots to the least observed landmarks, there might be landmarks with fewer observation count which may be far away
from the current frontier and $ISW$ tries to navigate the robots to them. Moreover, since $ISW$ will only increase the landmark's observation count by one unit, robots may need to visit the current
landmark again soon after they observed the next landmark. In other words, after observing the least observed landmark, the next step of $ISW$ will leave this obtained landmark for one that has been
observed less. Therefore, the consecutive steps of $ISW$ will cause robots jumping between different landmarks. On the other hand, since robots have directional sensors, few observations are needed to
capture the simplices which are connected to that landmark. In order to capture all the landmark's neighbors, robots need to approach to the corresponding landmark with different orientations. 

Another issue caused by having consecutive $ISW$ is that, the frontier of the simplicial complex could grow non-uniformly in one particular direction. Essentially, $ISW$ tries to grow the boundaries and it
always do that with the closest landmark. Consequently, after too many consecutive steps of $ISW$ the boundary will have a horn-like shape as depicted in Figure \ref{fig12}. In other words, $ISW$ is only
useful for acquiring new frontiers.

Hence, to solve both of these issues, few steps of $RW$ is needed to explore the adjacent landmarks to ensure they have been observed sufficient number of times before switching back to $ISW$.

\subsubsection{{Homology Informed Walk}}
Once the robots performed $RW$ and $ISW$, there might still remain some holes in the constructed landmark complex coverage $\mathcal{K}$ due to the insufficient number of observations. We call them
\emph{false holes} (holes in the Landmark complex due to insufficient exploration -- See Figure \ref{Simplicial_Complex}), as opposed to the holes generated in the landmark complex due to the presence of obstacles. In order to localize them and navigate the robots to
these \emph{false holes}, we used homology theory to detect $1$-simplices that bound them (See Figure \ref{Thresholding}). In the following subsections, we briefly describe how tools from homology theory can be used to
identify these $1$-simplices and navigate robots to them.
Fore more details on these methods the reader can refer to~\cite{Hatcher:AlgTop,jasons_paper}.

\textbf{\emph{i.} Boundary matrices and higher order Laplacian}:
In order to use the homology theory, first we need to define the boundary matrices $B_1$ and $B_2$ to be $n \times m$ and $m\times p$ matrices, where $n$, $m$, and $p$ respectively are the number of
$0$-simplices, $1$-simplices, and $2$-simplices.

\begin{figure}[h]
	\centering
	\includegraphics[width=2.5in]{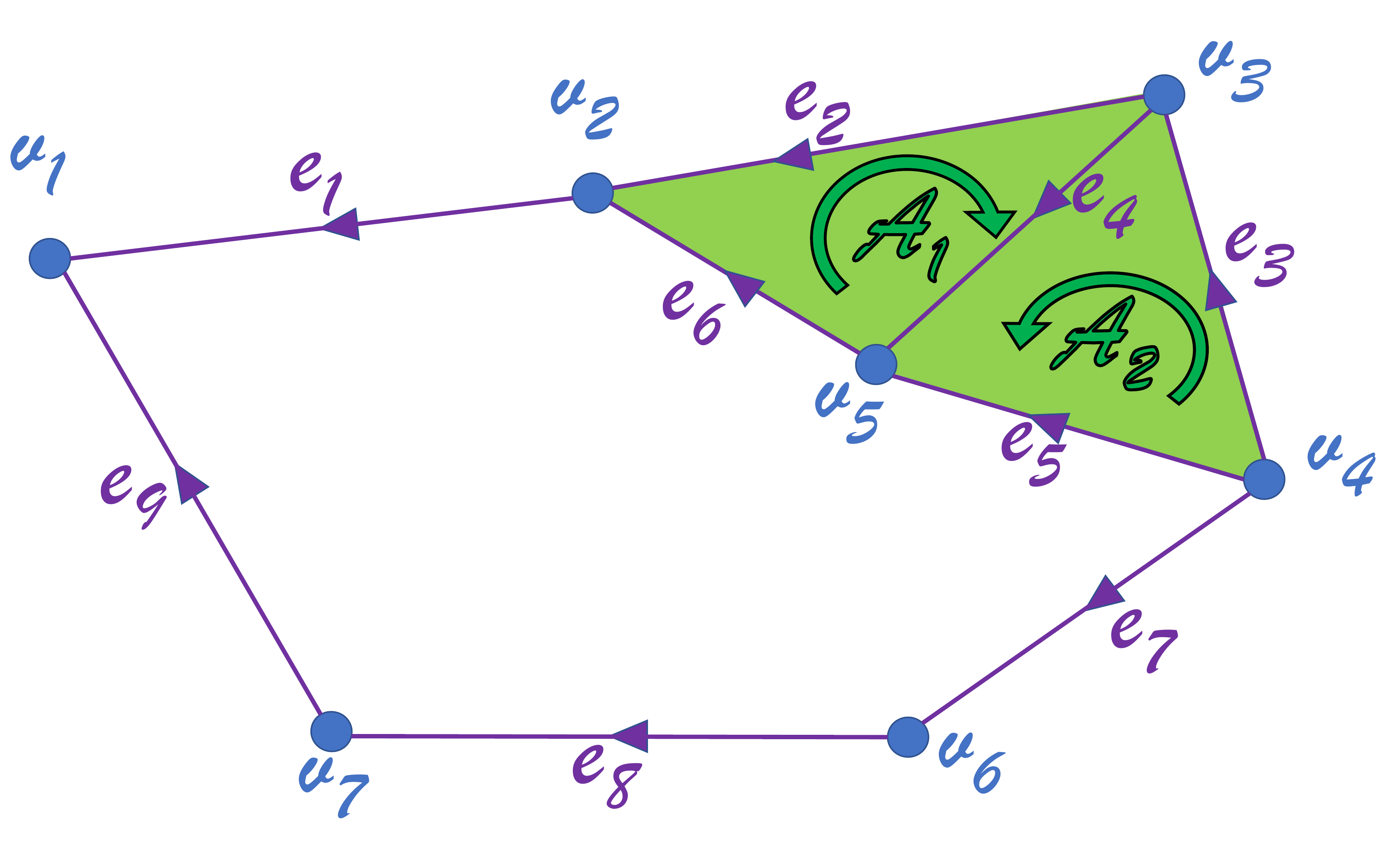}
	\caption{A directed simplicial complex of dimension 2.}
	\label{boundary_matrices}
\end{figure}

\begin{figure}
\centering
\subfloat[As shown, there is a false hole in the simplicial complex]{
\includegraphics[width=2.25in]{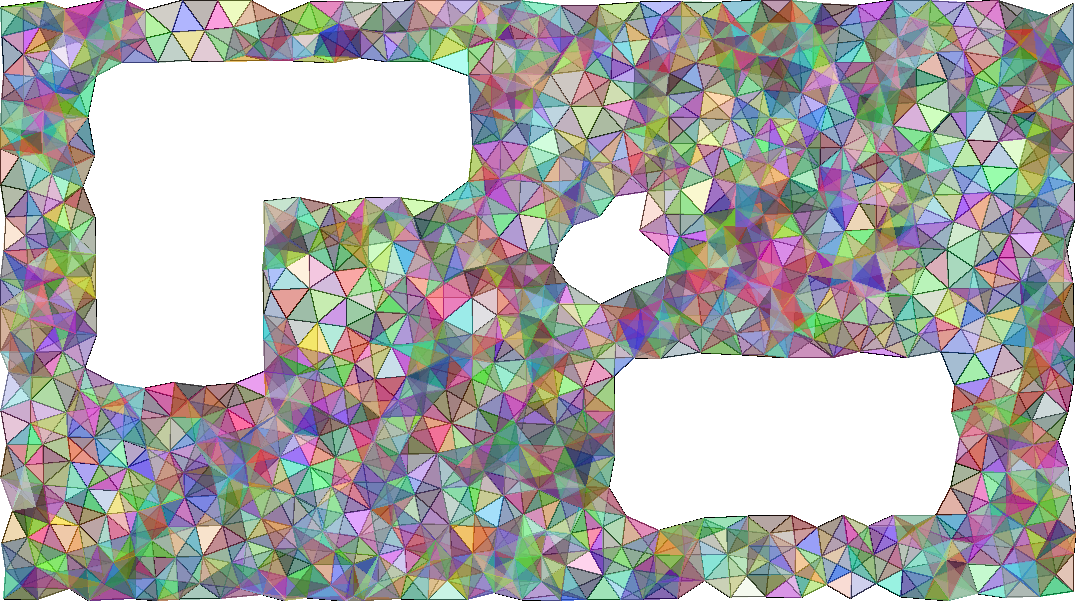}
\label{Simplicial_Complex}}\hfil
\subfloat[The corresponding result after solving the Laplacian Dynamics]{
\includegraphics[width=2.25in]{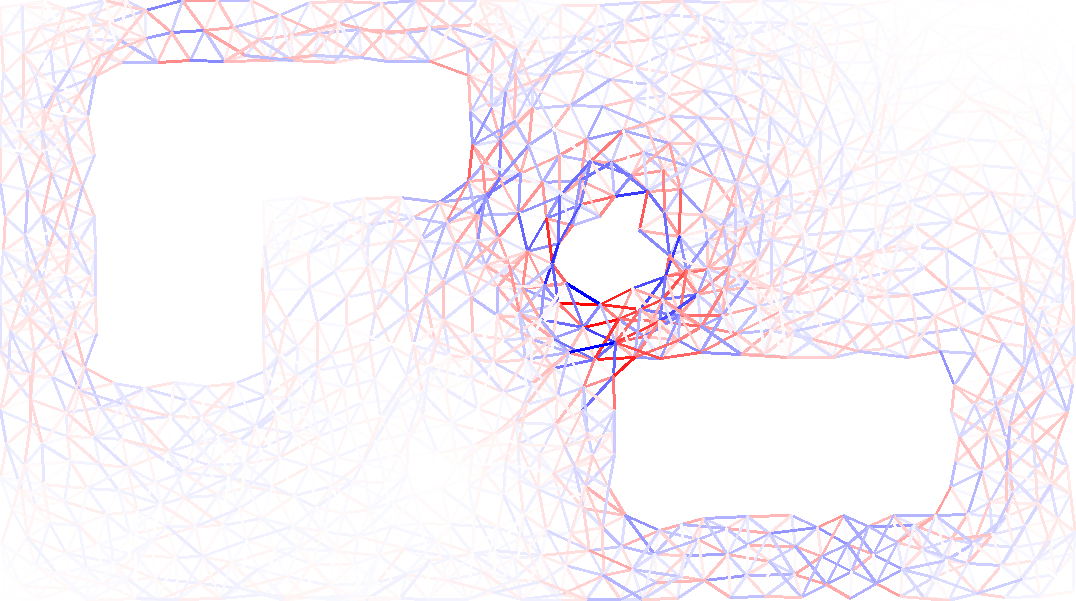}
\label{Laplacian_Dynamics}}\hfil
\subfloat[Result after $\ell_{1}$-Norm Minimization]{
\includegraphics[width=2.25in]{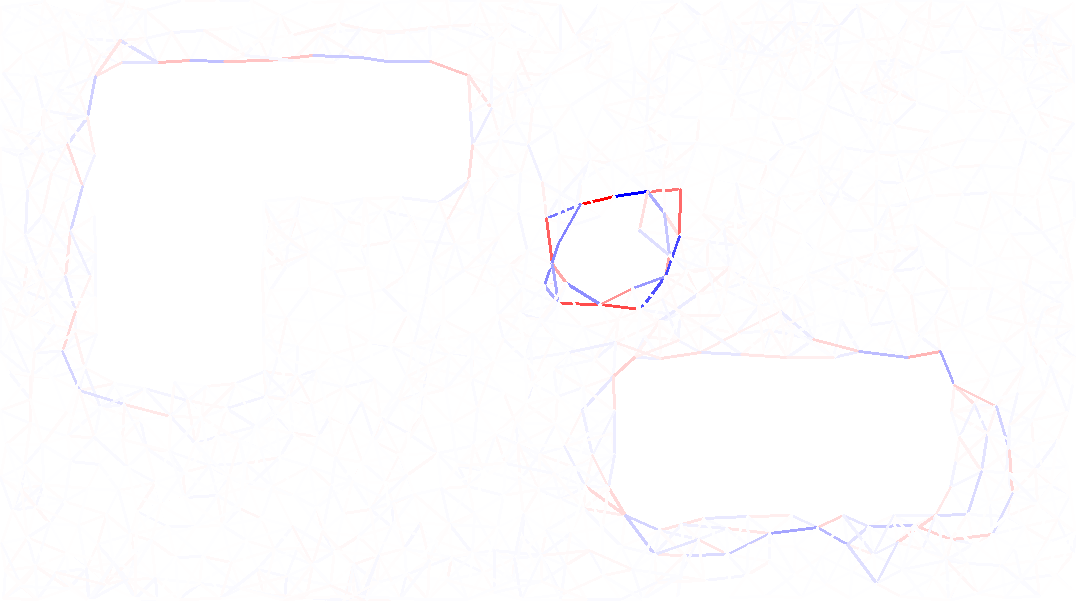}
\label{L1_norm_minimization}}\hfil
\subfloat[Result after choosing vertices that bound the false hole]{
\includegraphics[width=2.25in]{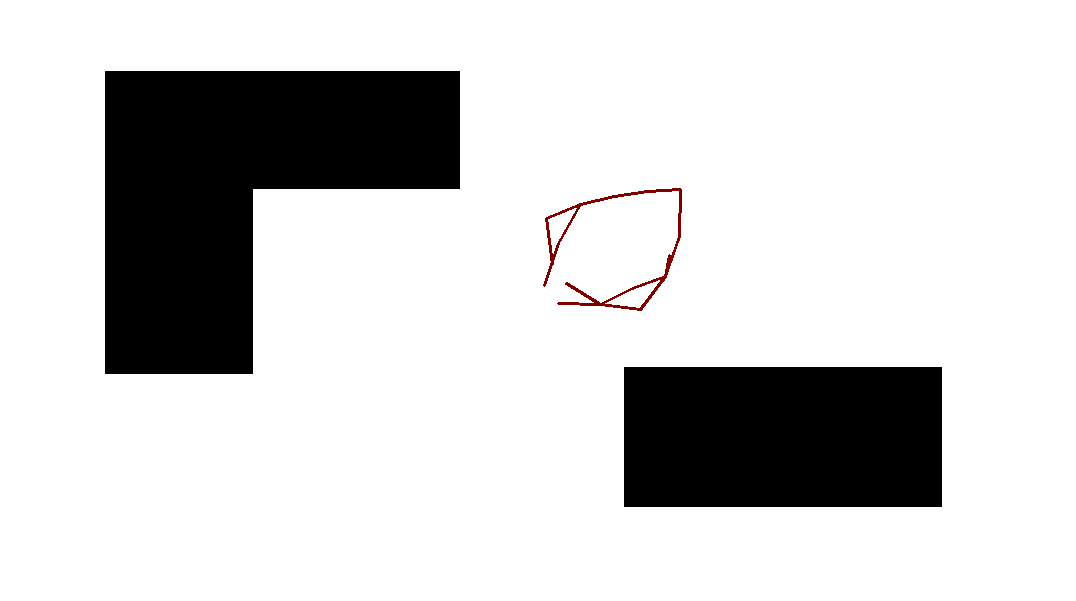}
\label{Thresholding}}\hfil
\caption{The Laplacian Dynamic and $\ell_{1}$-norm minimization can localize the holes in the simplicial complex}
\label{HIW_fig}
\end{figure}

Consider the simplicial complex in Figure \ref{boundary_matrices}. To construct $B_1$ and $B_2$ we need to assign arbitrary orientation to the $1$-simplices and $2$-simplices as shown in the figure. The
$(i,j)^{th}$ element of $B_1$ matrix, where $i$ is the identity of the vertex and $j$ is the identity of the $1$-simplex, can only have 0, +1, and -1 values. It will be 0 when the $i^{th}$ vertex is not
one of the ends of the $j^{th}$ $1$-simplex, +1 when $j^{th}$ $1$-simplex points towards the $i^{th}$ vertex, and -1 if $j^{th}$ $1$-simplex points away the $i^{th}$ vertex. For instance, in the given
example, the first column of $B_1$ which corresponds to $e_1$, will have +1 for the $(1,1)^{th}$ element, -1 for the $(2,1)^{th}$ element, and the rest will remain zero.

Similarly, the $(j,k)^{th}$ element of $B_2$, where $j$ is the identity of the $1$-simplex and $k$ is the identity of the $2$-simplex, also can only have 0, +1, and -1 values. It will be 0 when the
$j^{th}$ $1$-simplex is not adjacent to the $k^{th}$ $2$-simplex, +1 when they are adjacent and their orientations are aligned, and -1 if their orientation do not match. For instance the $(2,1)^{th}$
element of $B_2$ will be -1, whereas $(4,1)^{th}$ and $(6,1)^{th}$ elements of $B_2$ will have a +1 value while the rest is zero. $B_1$ and $B_2$ matrices of the simplicial complex given in Figure
\ref{boundary_matrices} are provided in the following.

\includegraphics[width=3.6in]{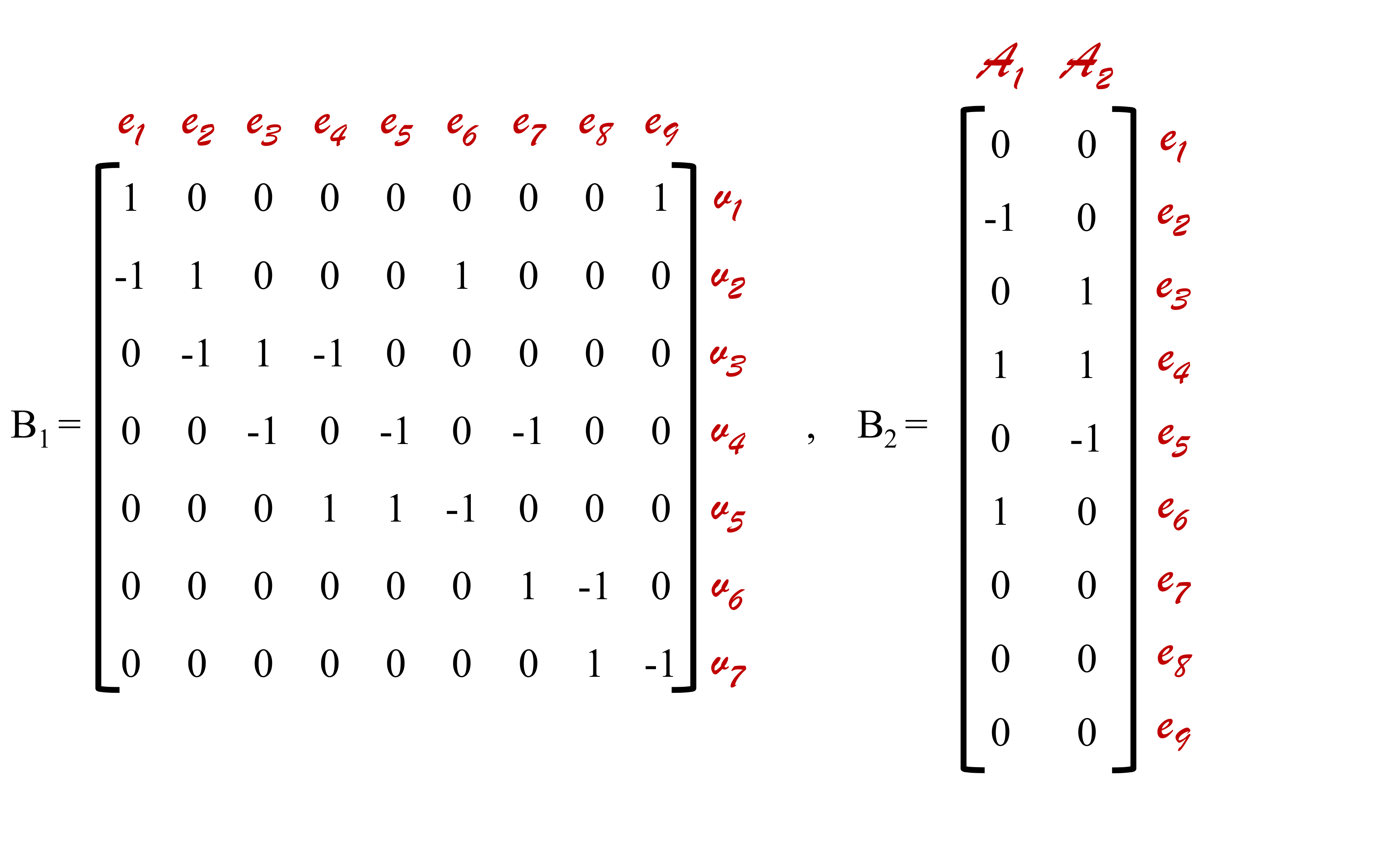}
% \includegraphics[width=1.50in]{boundary_matrix_B_2.pdf}
% \vspace{2mm}
% 
% $B_1$ = 
% $\left[\begin{smallmatrix}
% +1 & 0 & 0 & 0 & 0 & 0 & 0 & 0 & +1 \\
% -1 & +1 & 0 & 0 & 0 & +1 & 0 & 0 & 0 \\
% 0 & -1 & +1 & -1 & 0 & 0 & 0 & 0 & 0 \\
% 0 & 0 & -1 & 0 & -1 & 0 & -1 & 0 & 0 \\
% 0 & 0 & 0 & +1 & +1 & -1 & 0 & 0 & 0 \\
% 0 & 0 & 0 & 0 & 0 & 0 & +1 & -1 & 0 \\
% 0 & 0 & 0 & 0 & 0 & 0 & 0 & +1 & -1 \\
% \end{smallmatrix}\right]$
% 
% \vspace{2mm}
% 
% $B_2$ = 
% $\left[\begin{smallmatrix}
% 0 & 0 \\
% -1 & 0 \\
% 0 & +1 \\
% +1 & +1 \\
% 0 & -1 \\
% +1 & 0 \\
% 0 & 0 \\
% 0 & 0 \\
% 0 & 0 \\
% \end{smallmatrix}\right]$
% 
% \vspace{2mm}

It can be shown that the product of $B_1$ and $B_2$ is always equal to $0$.
Furthermore, the $1^\text{st}$ order Laplacian matrix $\mathcal{L}_1$ is defined as following.

\begin{equation}
\mathcal{L}_1 = B_{1}^TB_{1} + B_{2}B_{2}^T
\end{equation}
We are interested in a specific element from the kernel of $\mathcal{L}_1$ which corresponds to a set of $1$-simplices forming tight \emph{cycles} around the \emph{holes} in the complex.

\textbf{\emph{ii.} Laplacian dynamics and $\ell_{1}$-norm minimization}:
In order to identify the $1$-cycles that bound the \emph{false holes} tightly, we used the results of \cite{jasons_paper}. In this work, the authors considered the combinatorial Laplacian
flow

\begin{equation} \label{eq:laplacian}
\dot{x}(t) = -\mathcal{L}_1x(t) , x(0) \in \mathbb{R}^{m}
\end{equation}

Starting with any $x(0) \in \mathbb{R}^m$, where $m$ is the number of $1$-simplices in the landmark complex $\mathcal{K}$, it can be shown that the Laplacian flow will converge to $\ker \mathcal{L}_1$
where $x$ is a linear combination of $1$-cycles that bounds the $1$-dimensional holes in the complex (See Figure \ref{Laplacian_Dynamics}). In order to get the tightest $1$-cycle around a hole we need to
solve the following $\ell_{1}$-norm optimization problem.

\begin{equation}
\min_{z \in \mathbb{R}^{p}} \lVert x + B_2z \lVert_1
\end{equation}

To solve this problem, a subgradient method is used \cite{subgradient_method}.

\begin{equation} \label{eq:subgradient}
z^{k+1} = z^{k} - \alpha_{k}B_{2}^{T}sgn(B_2z^{k} + x)
\end{equation}

The initial condition used in equation (\ref{eq:subgradient}) is $z^0 = 0,~z \in \mathbb{R}^p$ where $p$ is the number of $2$-simplices in the landmark complex $\mathcal{K}$. Moreover, $\alpha_k$ is the
step size which by picking a small enough $\alpha_k$, the converged $z$ gets close to the optimal solution (See Figure \ref{L1_norm_minimization}).

\begin{figure}[t]
\centering
\subfloat[]{
\includegraphics[width=1.4in]{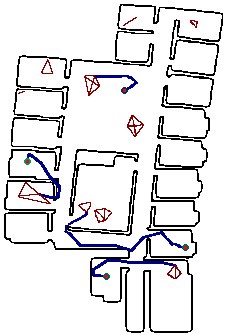}
\label{HungAlgo}}\hfil
\subfloat[]{
\includegraphics[width=1.4in]{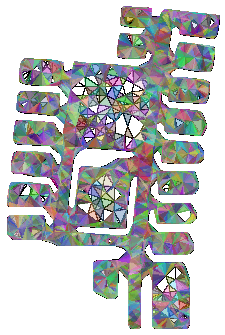}
\label{HungAlgo_simplex}}\hfil
\caption{Hungarian assignments with corresponding shortest paths to the false holes in the simplicial complex. In Figure (a) some of the false holes and the assigned robots with the shortest paths to each
assignment are shown, and Figure (b) shows the corresponding Landmark complex (note that the Landmark complex itself is an abstract simplical complex. We immerse it in $\mathbb{R}^2$ just for the purpose of visualization).}
\label{HungAlgofig}
\end{figure}

\textbf{\emph{iii.} \emph{HIW} algorithm and navigation}:
The pseudo-code given in Algorithm \ref{HIW_algo}, presents the outline of the $HIW$ algorithm and navigation. 

\begin{algorithm}
    \begin{algorithmic}[1]
        \caption{$Homology\_Informed\_Walk(\mathcal{R})$} \label{HIW_algo}
        \Require
        \Statex \textbf{a.} Landmark Complex $\mathcal{K}$ (Global Variable) with $n$ number of $0$-simplices, $m$ number of $1$-simplices, and $p$ number of $2$-simplices.
        \Statex \textbf{b.} Graph $G$ (with vertex set $\mathcal{V}(G)$, edge set $\mathcal{E}(G) \subseteq \mathcal{V}(G) \times V(G)$ and cost function $C_G$)
	\Statex \textbf{c.} Set of Agents location: $\mathcal{R} = \{r_i \in \mathcal{C} \subset SE(2) \mid 1 \leq i \leq N\}$
	\Statex \hrulefill
	\State Initiate $\mathcal{Q} = \{\}$\null\hfill\textcolor{gray}{//Set of $1$-simplices identities}  
	\State Initiate $\mathcal{C}c = \{\}$\hfill\textcolor{gray}{//Global variable of connected components' identity}
	\State $x^*:=$ Converged solution of $\dot{x}(t) = -\mathcal{L}_1x(t)$
	\State $z^*:=$ argmin$_{z \in \mathbb{R}^{p}} \lVert x + B_{2}z \lVert_1$
	\State Set $y = x^* + B_2z^*$
	\For {$i \in \{1, 2, \cdots, m\}$}\hfill\textcolor{gray}{//$m$ is the number of $1$-simplices in $\mathcal{K}$}
	  \If {$\lvert y[i] \rvert > \zeta$}\null\hfill\textcolor{gray}{//The $i^{th}$ element of vector $y$}
	    \State Set $\mathcal{Q} \leftarrow \mathcal{Q} \cup i$
	  \EndIf
	\EndFor
	\State Graph $G^\prime :=$ $Construct\_Graph(\mathcal{Q})$
	\State Set $\mathcal{C}c :=$ $Identify\_Connected\_Components(G^\prime)$\hfill\textcolor{gray}{//By using Dijkstra algorithm on graph $G^\prime$}
	\ForAll {$i \in \{1, 2, \cdots, N\}$ \textbf{on individual threads}}
	  \While {$\mathcal{C}c \neq \emptyset$}
	    \State $\mathcal{C}t :=$ $Compute\_Cost\_Matrix(\mathcal{R},~\mathcal{C}c,~G)$
	    \hfill\textcolor{gray}{//Cost matrix of size robots ($N$) times clusters ($M$)}
	    \State $j^* :=$ $Hungarian\_Assignment(\mathcal{C}t,~i)$
	    \For {$k \in \{1, 2, \cdots, w\}$}\hfill\textcolor{gray}{//$w$ is the number of $0$-simplices in $\mathcal{C}c[j^*]$}
	      \State $r_i \leftarrow Navigate(r_i,~\mathcal{C}c[j^*][k])$
	    \EndFor
	    \State Set $\mathcal{C}c = \mathcal{C}c - \mathcal{C}c[j^*]$\null\hfill\textcolor{gray}{//Remove $\mathcal{C}c[j^*]$ from $\mathcal{C}c$}
	  \EndWhile
	\EndFor
    \end{algorithmic}
\end{algorithm}

In line 3 and 4 of this pseudo-code, $x^*$ and $z^*$ respectively are the converged solutions of equations (\ref{eq:laplacian}) and (\ref{eq:subgradient}). In line 6 through 10, the algorithm checks the
absolute value of every element in vector $y = x^* + B_2z^*$ ($y \in \mathbb{R}^m$), and if it is greater than a computed threshold ($\zeta$), the identity of that $1$-simplex will be inserted to
$\mathcal{Q}$. The higher the value of $\lvert y[i] \rvert$ for the $i^{th}$ $1$-simplex, it is more likely to be adjacent to a hole. It is notable that $\zeta$ is computed using the standard deviation of
$\lvert y[i] \rvert, i = 1, 2, \cdots, m$.
\changed{
Since a large number of elements in vector $y$ have values close to zero and only a small fraction have absolute values greater than zero (See Figure \ref{L1_norm_minimization}), if we order the elements
in vector $y$ by absolute values, we would be able to see a jump. In order to find the appropriate thresholding value at where the jump occurs, we use the standard deviation of vector
$\lvert y[i] \rvert, i = 1, 2, \cdots, m$ to compute $\zeta$, independent from size of the vector $y$.

Since we selected the $1$-simplices with highest absolute value in vector $y$, these edges constitute the tightest $1$-cycle that bound holes in the landmark complex $\mathcal{K}$ (See Figure
\ref{Thresholding}). Furthermore, since these $1$-simplices constitute of connected components surrounding isolated holes, we need to identify each of them. Therefore, we construct graph $G^\prime$ such
that the vertex set $\mathcal{V}(G^\prime)$ and the edge set $\mathcal{E}(G^\prime)$ are the $0$-simplices and $1$-simplices in $\mathcal{Q}$ (Line 11). Afterwards in line 12, the function
$Identify\_Connected\_Components$ takes the graph $G^\prime$ as an input and by using Dijkstra algorithm on $G^\prime$, is able to identify these connected components. The output of this function is the
set $\mathcal{C}c$ where the $i^{th}$ element in the set itself is a set of $0$-simplices corresponding to the $i^{th}$ connected component.
}

In lines 13 to 20, each robot will find its own assignment to a connected component and will navigate to explore and observe the $0$-simplices in order to cover the holes. This is executed on an individual
thread for each robot. In order to assign the connected components to the robots, we used the Hungarian Algorithm, since it is of a cubic complexity \cite{HungAlgo}. To use Hungarian Algorithm, we
compute the cost matrix $\mathcal{C}t$ (line 15) which is of size $N \times M$, where $N$ is number of robots and $M$ is the number of connected components. The $(i,j)^{th}$ element of $\mathcal{C}t$
constitutes of the distance from the $i^{th}$ robot to the $j^{th}$ connected component and it is computed by running the Dijkstra algorithm on the graph $G$ constructed over the landmark complex
$\mathcal{K}$. Furthermore, in line 16, the $Hungarian\_Assignment$ function, takes the cost matrix $\mathcal{C}t$ and the identity of the robot running this thread as inputs, and returns the identity of
the assigned connected component ($j^*$). Lines 17 to 19 of Algorithm \ref{HIW_algo}, navigate the $i^{th}$ robot to each $0$-simplices in $\mathcal{C}c[j^*]$ using the function $Navigate(r_i,~i^*)$
described in Algorithm \ref{nav}.
By navigating each robot to these connected components, we explore the regions corresponding to the holes in the coverage, hence we are able to cover the false holes much faster than combined $RW$ and
$ISW$. In Figure \ref{HungAlgofig} the Hungarian assignment with the corresponding shortest path to the assigned connected component for each robot is depicted.

\begin{figure}
\centering
\subfloat[The constructed simplicial complex $\mathcal{K}^{t}$ after performing combined $RW$ and $ISW$]{
\includegraphics[width=0.18\textwidth]{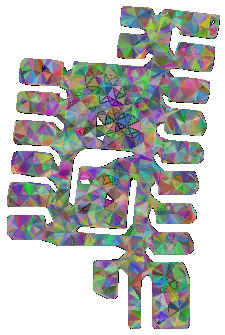}
\label{HIW_BC_101386_RWISW}} \hspace{0.01in}
\subfloat[Detected false holes in the simplicial complex $\mathcal{K}^{t}$]{
\includegraphics[width=0.18\textwidth]{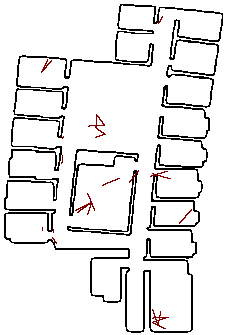}
\label{HIW_BC_first_homology}}\hspace{0.01in}
\subfloat[Fixed simplicial complex $\mathcal{K}^{t+1}$ after performing the first step of $HIW$]{
\includegraphics[width=0.18\textwidth]{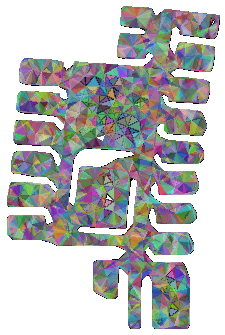}
\label{HIW_BC_110870_HIW}}\hspace{0.01in} \newline
\subfloat[Detected false holes in the simplicial complex $\mathcal{K}^{t+1}$]{
\includegraphics[width=0.18\textwidth]{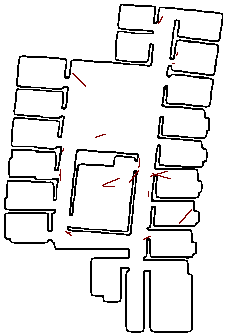}
\label{HIW_BC_second_homology}}\hspace{0.01in}
\subfloat[Final Landmark complex of the environment after performing the second step of $HIW$]{
\includegraphics[width=0.18\textwidth]{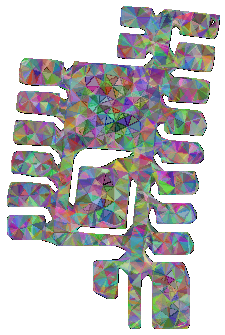}
\label{BC_final_HIW}}\hfil
\caption{Detection and exploration of \emph{false holes} using $HIW$ (note that the Landmark complex itself is an abstract simplical complex. We immerse it in $\mathbb{R}^2$ just for the purpose of visualization)}
\label{fig14}
\end{figure}

\changed{Results of the $HIW$ algorithm is presented in Figure \ref{fig14}. In Figure \ref{HIW_BC_101386_RWISW}, 4 robots have performed combined $RW$ and $ISW$ to construct the simplicial complex.
Afterwards, robots switch to $HIW$ to detect the holes in the simplicial complex. We explain the switching strategy in detail in Section~\ref{sec:results}, where we experimentally discover the optimal criteria for switching. We assume that robots are able to
detect whether a landmark is adjacent to an obstacle or not. Therefore by using this information robots are able to distinguish the false holes from the holes that are corresponding to the obstacles in the
environment.
}

\subsection{Landmark Complex Construction Algorithm (LCCA)} \label{sec:LCCA}
\changed{
So far we have explained necessary concepts and subroutines for the overall Landmark Complex Construction Algorithm ($LCCA$). In this section we put these concepts together and develop Algorithm \ref{LCCA} for enabling a group of robots to create a Landmark complex representation of an environment through a balanced strategy of exploration and exploitation. At each time step, each robot will make an observation to create a simplex from the observed landmarks and to update the
landmark complex $\mathcal{K}$. Moreover, this algorithm runs for each robot on a separate thread (Line 1 through 21).

Since at the beginning of the exploration, the location of landmarks and robots are unknown, robots will perform a fixed number of steps ($\gamma$) of \emph{RW}, to construct a partial landmark complex in
order to localize themselves with respect to the landmarks (Lines 2 through 4). Once robots have created a sufficiently large landmark complex, they switch to the combined \emph{RW} and \emph{ISW} (Lines 6
through 20) until the growth rate of landmark complex $\mathcal{K}$ become less than $\varepsilon_1$. Furthermore, \emph{RW} takes the location of the $i^{th}$ robot running the thread as an input ($r_i$),
where \emph{ISW} will take the set of all robots' location $\mathcal{R}$ and the identity of the $i^{th}$ robot to find the goal landmark and the distance to it for the $i^{th}$ robot. However, \emph{ISW}
also uses the established landmark complex $\mathcal{K}$, which is a global variable, to construct graph $G$ for Voronoi partitioning algorithm, described in Algorithm \ref{VP_algo}, and to find the
$g\_score$ to the goal landmark.

In addition, \emph{LCCA} interleaves between \emph{ISW} and \emph{RW} to avoid issues caused by too many consecutive steps of \emph{ISW}. As described in \emph{Informed Systematic Walk} subsection, these
issues are, first, navigating the robots to the far frontier landmarks back and forth, and second, growing the landmark complex non-uniformly. To this end, in line 7, the algorithm checks whether the
$i^{th}$ robot is at the boundary of the landmark complex or not, by checking the $g\_score$ of the acquired goal landmark. Also it checks the number of which \emph{ISW} has been performed consecutively.
If the $g\_score$ is less than a fixed number $\omega$, or if the counter on consecutive steps of \emph{ISW} is greater than a fixed number $\eta$, \emph{LCCA} will execute $\delta$ number of \emph{RW}
(Lines 15 through 19). 

At the final stages of the exploration, \emph{LCCA} will switch to \emph{Homology Informed Walk}, explained in Algorithm \ref{HIW_algo}. This will enable the robots to locate holes in the landmark complex
$\mathcal{K}$ and to cover the false holes much faster than \emph{ISW} and \emph{RW}. When the growth rate of the landmark complex $\mathcal{K}$ becomes less than $\varepsilon_2$, where
$\varepsilon_2 < \varepsilon_1$, the exploration will be stopped.
}

\begin{algorithm}
    \begin{algorithmic}[1]
        \caption{Landmark Complex Construction Algorithm} \label{LCCA}
        \Require
        \Statex \textbf{a.} Set of Agents location: \Statex$\mathcal{R} = \{r_i \in \mathcal{C} \subset SE(2) \mid 1 \leq i \leq N\}$
        \Ensure
        \Statex \textbf{a.} Landmark Complex $\mathcal{K}$
        \Statex \hrulefill
	\For {$i \in \{1, 2, 3, \cdots, N\}$ \textbf{on individual thread}}
	  \For {$j \in (1, 2, \cdots, \gamma)$}
	    \State $r_i \leftarrow RW\_Observe(r_i)$
	  \EndFor
	  \While {rate of growth $\mathcal{K} \leq \varepsilon_1$}
	    \State $[i^*,~g\_score] := ISW(\mathcal{R},~i)$
	    \If {$~g\_score \leq \omega~$ \textbf{or} $~ISW\_counter \geq \eta~$}
	      \For {$k \in (1, \cdots, \delta)$}
		\State $r_i \leftarrow RW\_Observe(r_i)$
	      \EndFor
	      \State $ISW\_counter$ = 0
	    \Else
	      \State $r_i \leftarrow Navigate(r_i,~i^*)$
	      \State $ISW\_counter$++
	    \EndIf

	  \EndWhile
	\EndFor
	\While {rate of growth $\mathcal{K} \leq \varepsilon_2$}
	  \State $Homology\_Informed\_Walk(\mathcal{R})$
	\EndWhile
	\State \Return $\mathcal{K}$
    \end{algorithmic}
\end{algorithm}

\section{Results} \label{sec:results}
\changed{
Our proposed Landmark Placement Algorithm ($LPA$) described in Section~\ref{sec:landmark-placement} was implemented in C++.
The algorithm allows us to design strategic placement of landmarks in an environment during its design/construction in order to ensure that at least one landmark is visible to a robot sensor for every configuration
in $\mathcal{C} \subset {\mathbb{R}^2}$ (for disk-shaped sensor footprints) or $\mathcal{C} \subset SE(2)$ (for directional sensors). In Figures \ref{Landmark Placement Algorithm} and \ref{fig6}, results for populating a simple environment with landmarks are
provided for $\mathcal{C} \subset {\mathbb{R}^2}$ and $\mathcal{C} \subset SE(2)$ respectively. Moreover, Figure \ref{LPA_results} shows the results of Landmark Placement Algorithm ($LPA$) on two different
complex environments. These environments are populated with the landmarks where $\mathcal{C} \subset SE(2)$. In this figure, the \u{C}ech complexes of the visibility domain of the landmarks for these
environments are also depicted.

%In this paper, we proposed a metric-free method for exploration of the swarm of decentralized robots in an unknown environment 
We also implemented the Landmark Complex Construction Algorithm (LCCA) described in Section~\ref{sec:exploration-navigation}
using \emph{C++}. Our implementation is a multi-threaded one, with each thread emulating a robot, and with limited inter-thread communication to construct and update the Landmark complex maintained in the cloud (a central server).
We evaluated the proposed algorithm using two different complex environments refereed to as the \emph{first complex environment} and the \emph{second complex environment}.
Results of the Landmark Complex Construction Algorithm is presented in Figures \ref{L457-RW_ISW&HIW} and \ref{BC-RW_ISW&HIW} for the first and second complex environments respectively.
%
%To
%this end, first we proposed the short-term trajectory ($STT$) model based on Dubins curves for non-holonomic robots, such that with three variables $(\rho,~s,~\beta)$ robots can take a curve of radius
%$\rho$ and travel the distance of $s$ along that curve. The two-state variable $\beta$ allows robot to take the left or right curve to navigate towards the detected landmark. Afterwards, we proposed three
%different modes of walk for different circumstances. First, by performing Random Walk ($RW$), robots will be able to obtain a relative understanding of their initial localization with respect to the
%landmark and construct a partial simplicial complex. In the $RW$ all three variables of $STT$ are chosen randomly. Subsequently, in order to expand the simplicial complex, robots perform Informed
%Systematic Walk ($ISW$) to explore those landmarks that have been observed fewer times. And eventually, Homology Informed Walk will eliminate any false holes that appears in the simplicial complex that do
%not correspond to any obstacle in the environment.

\begin{figure}
	\centering
	\subfloat[observations count: 3000. Beginning of the combined $RW$ and $ISW$]
	{\includegraphics[width=1.50in]{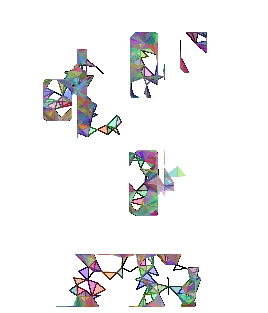}
		\label{3351_RW_ISW}}\hfil
	\subfloat[observations count: 7000]
	{\includegraphics[width=1.50in]{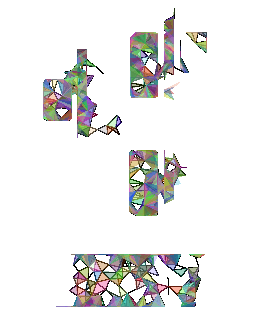}
		\label{7011_RW_ISW}}\hfil
	\subfloat[observations count: 10000]
	{\includegraphics[width=1.50in]{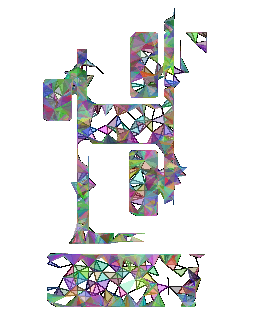}
		\label{10337_RW_ISW}}\hfil
	\subfloat[observations count: 20000]
	{\includegraphics[width=1.50in]{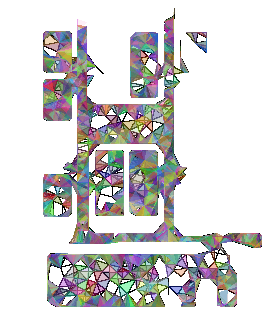}
		\label{20869_RW_ISW}}\hfil
	\subfloat[observations count: 30000]
	{\includegraphics[width=1.50in]{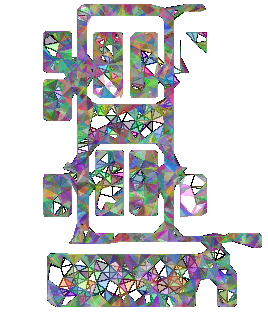}
		\label{30320_RW_ISW}}\hfil
	\subfloat[observations count: 50000.]
	{\includegraphics[width=1.50in]{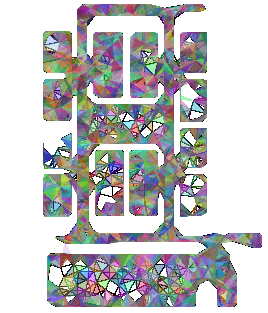}
		\label{51432_RW_ISW}}\hfil
	\subfloat[observations count: 100000. Final result after combined $RW$ and $ISW$.]
	{\includegraphics[width=1.50in]{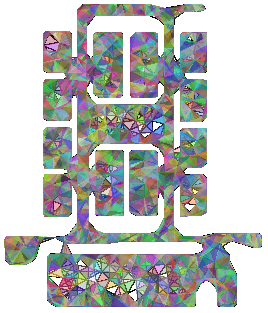}
		\label{98596_RW_ISW}}\hfil
	\subfloat[observations count: 180000. Final result after completing the $HIW$.]
	{\includegraphics[width=1.50in]{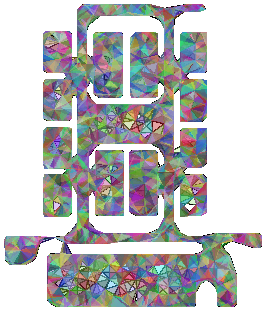}
		\label{187113_HIW}}\hfil
	\caption{LCCA in first complex environment (note that the Landmark complex itself is an abstract simplical complex. We immerse it in $\mathbb{R}^2$ and superimpose that on top of a map of the actual environment just for the purpose of visualization).}
	\label{L457-RW_ISW&HIW}
\end{figure}

\begin{figure}
	\centering
	\subfloat[observations count: 2500. Beginning of the combined $RW$ and $ISW$]
	{\includegraphics[width=1.12in]{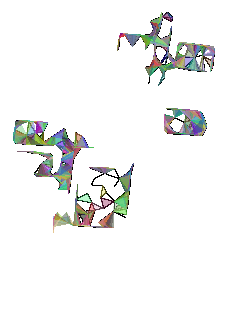}
		\label{BC_2469_RW_ISW}}\hfil
	\subfloat[observations count: 10000]
	{\includegraphics[width=1.12in]{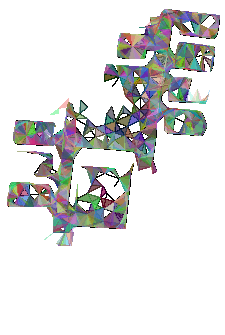}
		\label{BC_10828_RW_ISW}}\hfil
	\subfloat[observations count: 20000]
	{\includegraphics[width=1.12in]{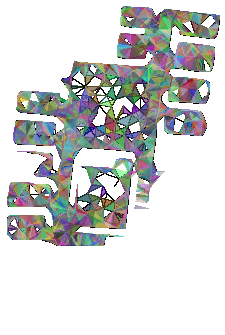}
		\label{BC_20434_RW_ISW}}\hfil
	\subfloat[observations count: 40000]
	{\includegraphics[width=1.12in]{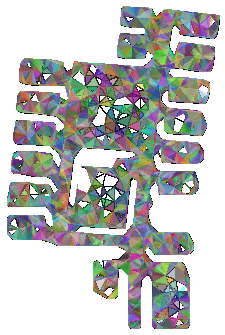}
		\label{BC_39922_RW_ISW}}\hfil
	\subfloat[observations count: 60000]
	{\includegraphics[width=1.12in]{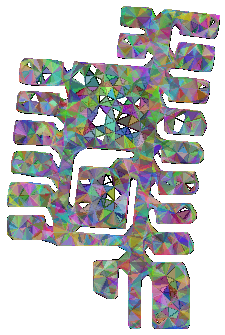}
		\label{BC_60281_RW_ISW}}\hfil
	\subfloat[observations count: 80000. Final results after exploring 72\% of the $2$-simplices by performing combined $RW$ and $ISW$]
	{\includegraphics[width=1.12in]{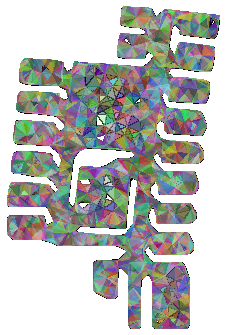}
		\label{BC_79939_RW_ISW}}\hfil
	\subfloat[observations count: 100000. Beginning of the $HIW$]
	{\includegraphics[width=1.12in]{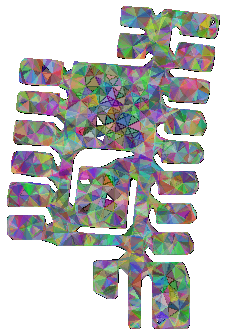}
		\label{BC_101386_RWISW}}\hfil
	\subfloat[observations count: 123500. Final result after completing the $HIW$.]
	{\includegraphics[width=1.12in]{BC_123655_HIW}
		\label{BC_123655_HIW}}\hfil
	\caption{LCCA in second complex environment (note that the Landmark complex itself is an abstract simplical complex. We immerse it in $\mathbb{R}^2$ and superimpose that on top of a map of the actual environment just for the purpose of visualization).}
	\label{BC-RW_ISW&HIW}
\end{figure}

\begin{figure}[h]
\centering
\includegraphics[width=3.25in]{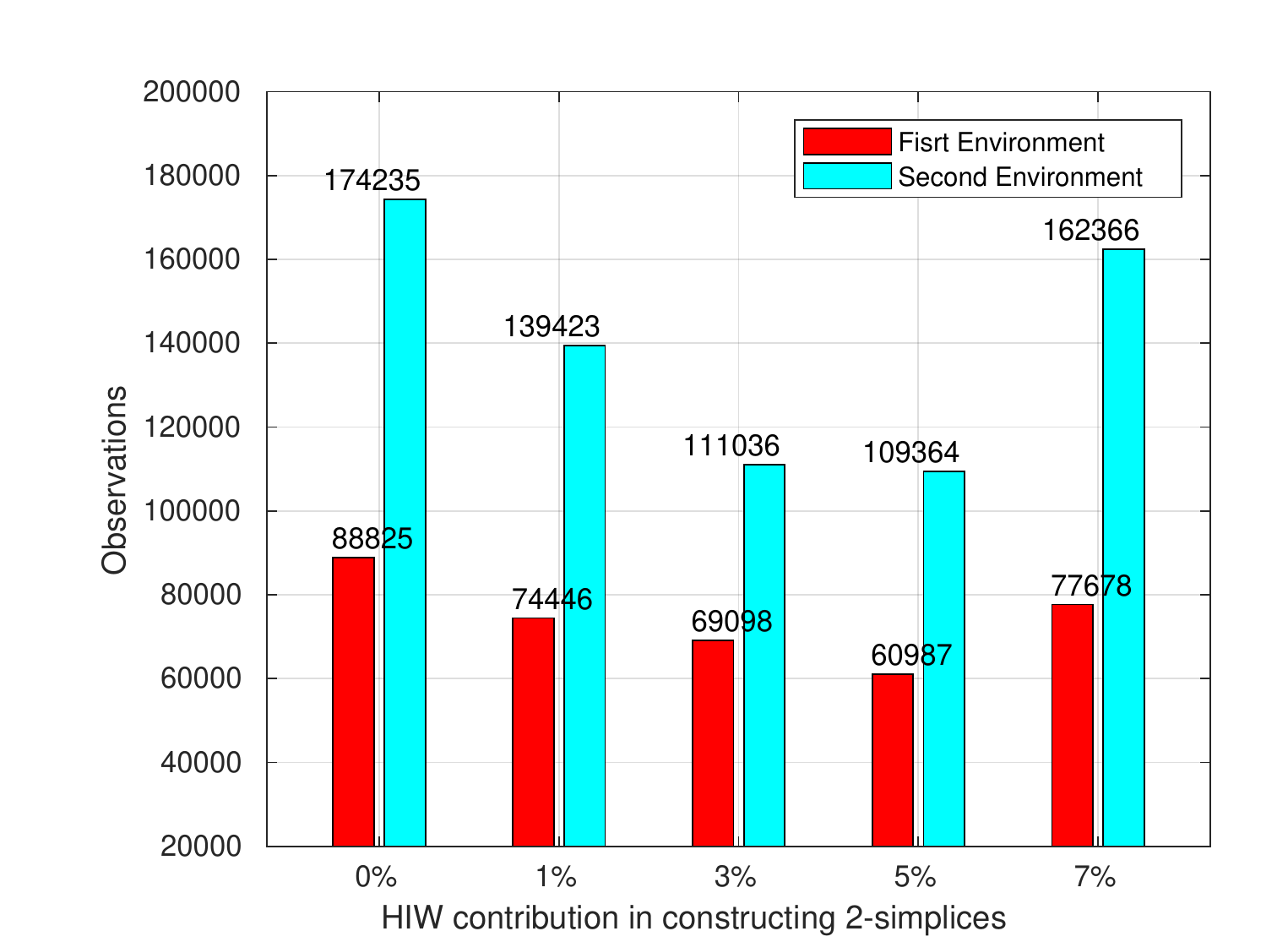}
\caption{Different $HIW$ contributions on constructing the $2$-simplices in the \u{C}ech complexes of the visibility domain of the landmarks, to reach the target amount for the first and second complex
environments.}
\label{Percentage_HIW}
\end{figure}

To evaluate the role of $HIW$ in improving exploration of the environment, we 
executed the LCCA with varying amounts of HIW and recorded the number of
%tested our method to construct certain number of simplices in the simplicial complex (both with $HIW$ and without it), and check how many
observations required.
As a reference/benchmark, we first counted all the $2$-simplices in the \u{C}ech complex of the visibility domain of the landmarks. Then we ran LCCA to construct 98\% of those $2$-simplices in the simplicial complex. 
We ran $5$ simulations for each of the complex environments, in which 4 robots performed combined $RW$ and $ISW$ to construct $98\%, 97\%,95\%,93\%$ and $91\%$ of the simplices respectively, following which $HIW$ was used to complete the remaining $0\%,1\%,3\%,5\%$ and $7\%$ respectively of the simplices to reach the target $98\%$.
%In the first simulation, 4 robots performed combined $RW$ and $ISW$ to reach the target amount (98\% of total $2$-simplices), and we recorded the number of
%total observations that all robots made. 
%In the next simulations we reduced the portion of combined $RW$ and $ISW$ and tried to reach the 98\% target with $HIW$. 
Figure \ref{Percentage_HIW} shows the
results of this experiment. We can deduce that $HIW$ improves the algorithm by reducing the total number of observations needed, to complete 98\% of the total $2$-simplices, however its effectiveness starts diminishing beyond a certain percentage.
In fact Figure \ref{Percentage_HIW} suggests that the
optimum level of $HIW$ use is to construct the last $5\%$ of total $2$-simplices with $HIW$.

Since the total number of $2$-simplices is unknown to the robots, we need a strategy for switching to $HIW$ without using this information. A useful data that robots have, is the \emph{growth rate of the
simplicial complex}. We define the growth rate of the simplicial complex ($r$), equal to the number of the new $2$-simplices added to the simplicial complex at each iteration divided by the total number of
$2$-simplices that are already existing in the simplicial complex. Figure \ref{ratio} shows the growth rate of the simplicial complex with time. As the exploration continues, it decreases exponentially
indicating that the number of new $2$-simplices added to the simplicial complex decreases with time.
We experimentally find the optimal value of $r$ at which the switch to $HIW$ results in completion of the exploration
with the minimum number of observations. To this end, we ran hundreds of simulations for each environment and recorded the number of total observations, as well as the growth rate value that the $HIW$ switch happened.
The results of these simulations are given in Figure \ref{ratio_BC_L457}. The x-axis denotes the growth rate value at which the switch happened and the y-axis is the total number of observations that robots made to complete the
exploration. It is noteworthy that switching on smaller $r$ corresponds to smaller contribution of $HIW$ in constructing the simplicial complex. While some $HIW$ does help with completing exploration with fewer observation, too much of $HIW$ increases the number of observations. This is expected since the purpose of $HWI$ is to fill \emph{holes} within the complex, and not complete exploration when there are unexplored frontiers. In fact, as evident from the plots in Figure~\ref{ratio_BC_L457}, a growth rate of about $0.004$ is the optimal value at which switching to $HIW$ minimizes the total observation count.

\begin{figure}
\centering
\includegraphics[width=3.25in]{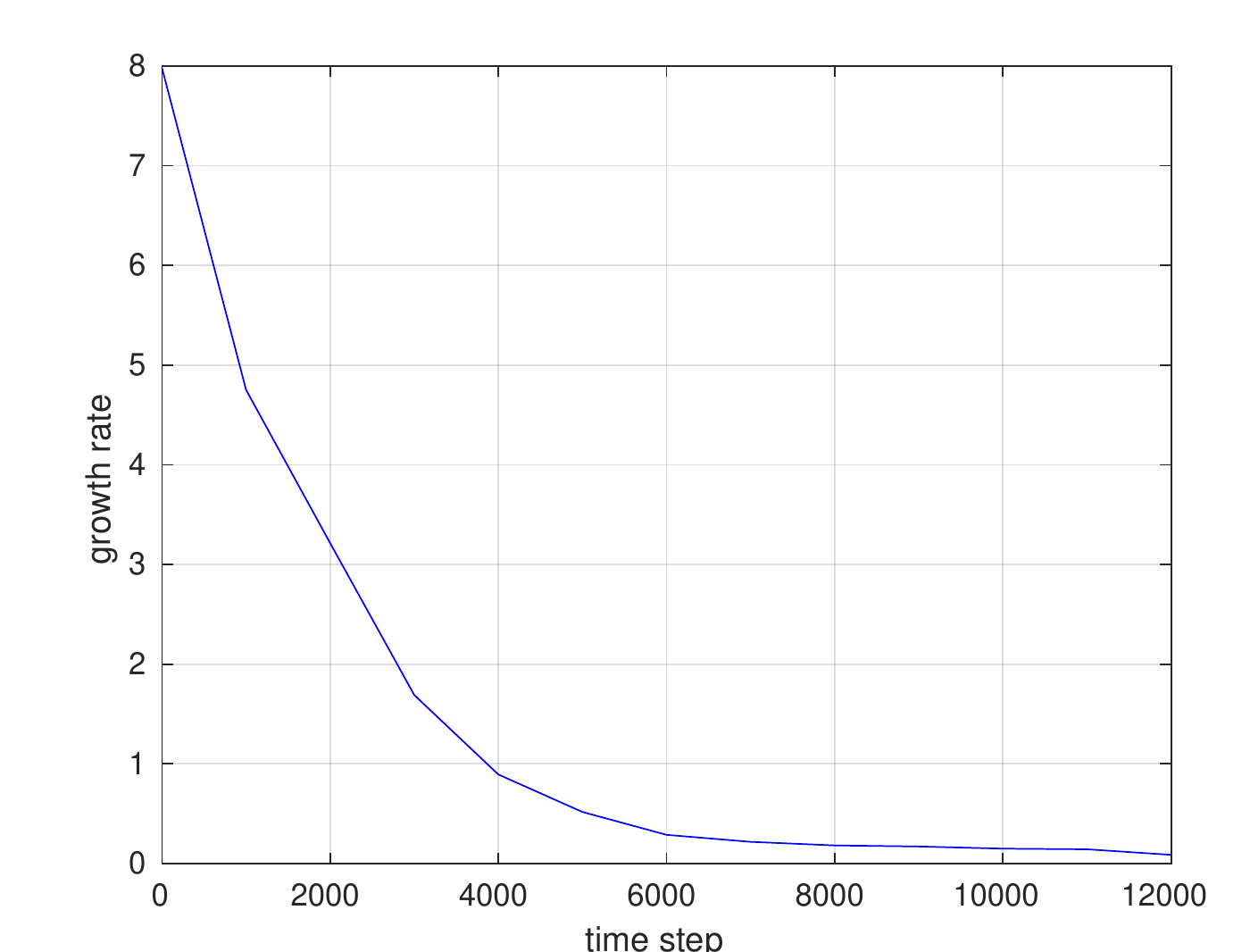}
\caption{Growth rate of the simplicial complex ($r$).}
\label{ratio}
\end{figure}

\begin{figure}
\centering
\subfloat[]
{\includegraphics[width=3.25in]{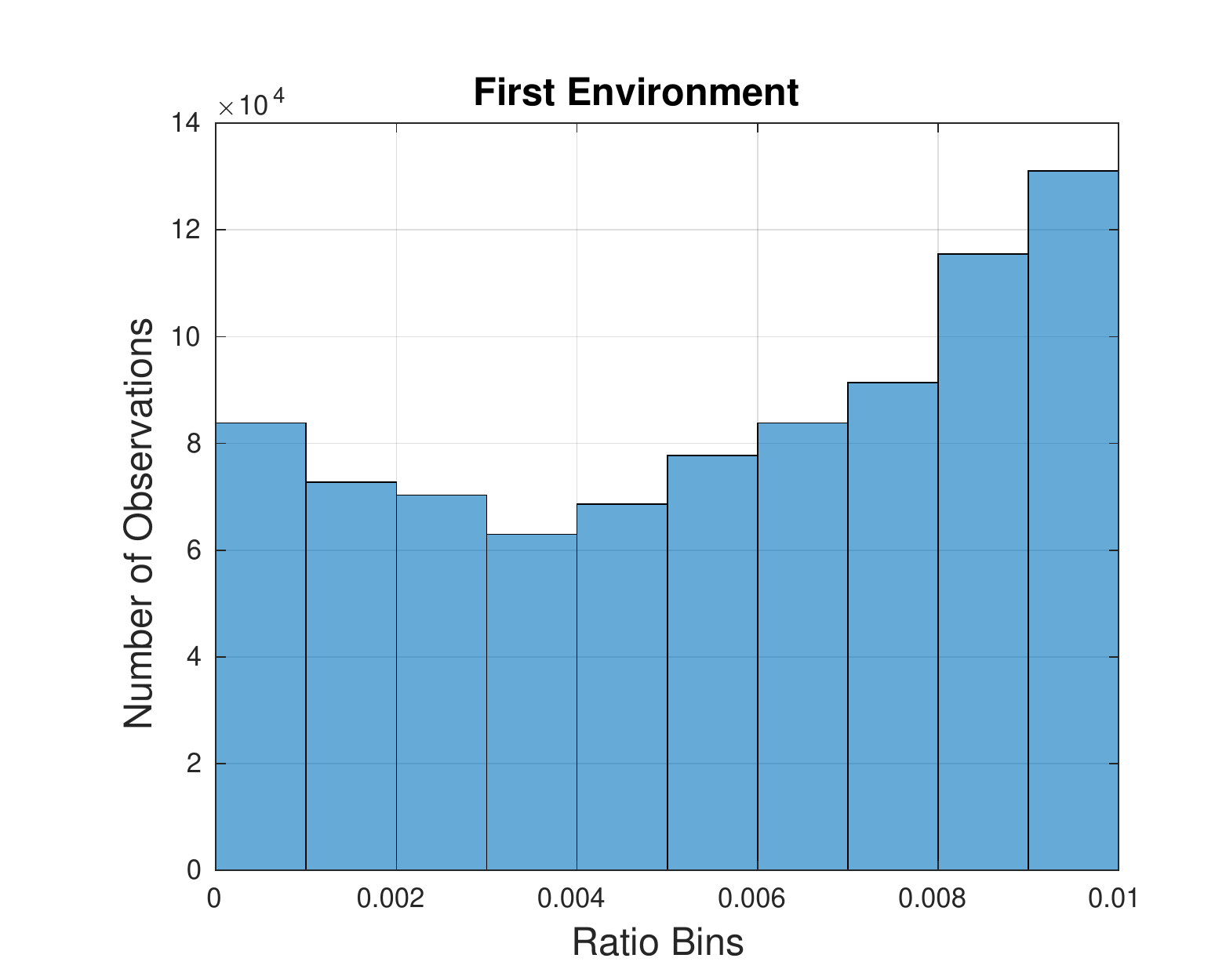}
\label{ratioL457}}
\subfloat[]
{\includegraphics[width=3.25in]{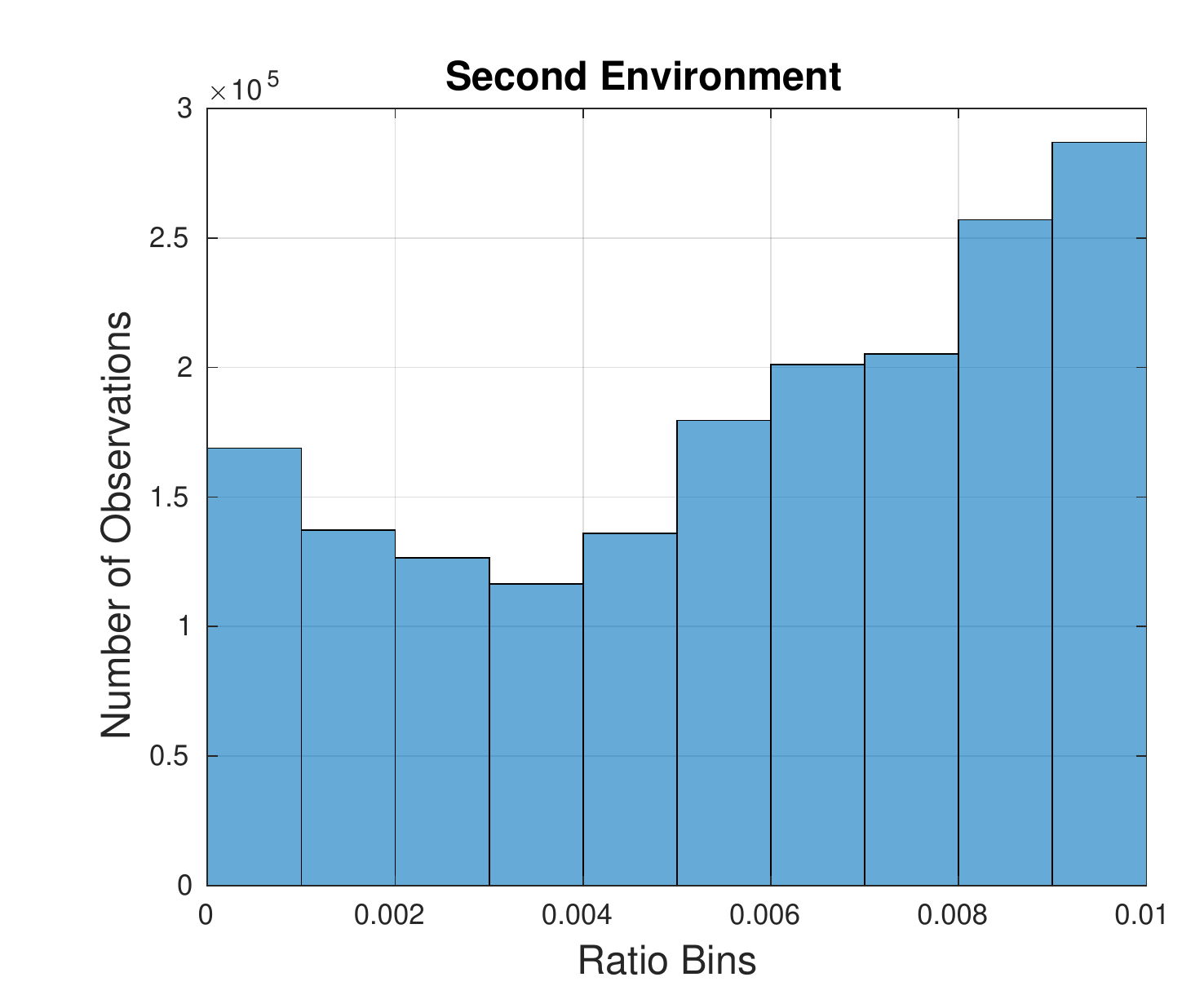}
\label{ratioBC}}\hfil
\caption{The number of the observations required to complete the simplicial complex plotted against the growth rate vale at which the switch to $HIW$ is performed. As evident, a switch at the growth rate of about $0.004$ gives an optimal performance in either of the environments.}
\label{ratio_BC_L457}
\end{figure}
}

\bibliographystyle{plain}
\bibliography{reference,references,s_star,thesis,markov1,slam,proposal,persistance_planning,new}

\begin{thebibliography}{10}

\bibitem{angeli2008fast}
Adrien Angeli, David Filliat, St{\'e}phane Doncieux, and Jean-Arcady Meyer.
\newblock Fast and incremental method for loop-closure detection using bags of
  visual words.
\newblock {\em IEEE Transactions on Robotics}, 24(5):1027--1037, 2008.

\bibitem{coverage:riemannian:IJRR:13}
Subhrajit Bhattacharya, Robert Ghrist, and Vijay Kumar.
\newblock Multi-robot coverage and exploration on riemannian manifolds with
  boundary.
\newblock {\em International Journal of Robotics Research}, 33(1):113--137,
  January 2014.
\newblock DOI: 10.1177/0278364913507324.

\bibitem{castellanos1999spmap}
Jose~A Castellanos, JMM Montiel, Jos{\'e} Neira, and Juan~D Tard{\'o}s.
\newblock The spmap: A probabilistic framework for simultaneous localization
  and map building.
\newblock {\em IEEE Transactions on Robotics and Automation}, 15(5):948--952,
  1999.

\bibitem{choset2001topological}
Howie Choset and Keiji Nagatani.
\newblock Topological simultaneous localization and mapping (slam): toward
  exact localization without explicit localization.
\newblock {\em IEEE Transactions on Robotics and Automation}, 17(2):125--137,
  2001.

\bibitem{davison2007monoslam}
Andrew~J Davison, Ian~D Reid, Nicholas~D Molton, and Olivier Stasse.
\newblock Monoslam: Real-time single camera slam.
\newblock {\em IEEE transactions on pattern analysis and machine intelligence},
  29(6):1052--1067, 2007.

\bibitem{jasons_paper}
J.~{Derenick}, V.~{Kumar}, and A.~{Jadbabaie}.
\newblock Towards simplicial coverage repair for mobile robot teams.
\newblock In {\em 2010 IEEE International Conference on Robotics and
  Automation}, pages 5472--5477, 2010.

\bibitem{dirafzoon_lobaton_2013}
A.~Dirafzoon and E.~Lobaton.
\newblock Topological mapping of unknown environments using an unlocalized
  robotic swarm.
\newblock In {\em 2013 ieee/rsj international conference on intelligent robots
  and systems (iros)}, page 5545–51, 2013.

\bibitem{Dubins1957OnCO}
L.~Dubins.
\newblock On curves of minimal length with a constraint on average curvature,
  and with prescribed initial and terminal positions and tangents.
\newblock {\em American Journal of Mathematics}, 79:497, 1957.

\bibitem{durrant2006simultaneous}
Hugh Durrant-Whyte and Tim Bailey.
\newblock Simultaneous localization and mapping: part i.
\newblock {\em IEEE Robotics \& Automation Magazine}, 13(2):99--110, 2006.

\bibitem{10.1007-978-3-319-10605-2_54}
Jakob Engel, Thomas Sch{\"o}ps, and Daniel Cremers.
\newblock Lsd-slam: Large-scale direct monocular slam.
\newblock In David Fleet, Tomas Pajdla, Bernt Schiele, and Tinne Tuytelaars,
  editors, {\em Computer Vision -- ECCV 2014}, pages 834--849, Cham, 2014.
  Springer International Publishing.

\bibitem{engel2015large}
Jakob Engel, J{\"o}rg St{\"u}ckler, and Daniel Cremers.
\newblock Large-scale direct slam with stereo cameras.
\newblock In {\em 2015 IEEE/RSJ International Conference on Intelligent Robots
  and Systems (IROS)}, pages 1935--1942. IEEE.

\bibitem{forster2014svo}
Christian Forster, Matia Pizzoli, and Davide Scaramuzza.
\newblock Svo: Fast semi-direct monocular visual odometry.
\newblock In {\em 2014 IEEE international conference on robotics and automation
  (ICRA)}, pages 15--22. IEEE, 2014.

\bibitem{ghrist2012topological}
R~Ghrist, D~Lipsky, J~Derenick, and A~Speranzon.
\newblock Topological landmark-based navigation and mapping.
\newblock {\em University of Pennsylvania, Department of Mathematics, Tech.
  Rep}, 8, 2012.

\bibitem{Hatcher:AlgTop}
A.~Hatcher.
\newblock {\em Algebraic Topology}.
\newblock Cambridge Univ. Press, 2001.

\bibitem{howard2006multi}
Andrew Howard.
\newblock Multi-robot simultaneous localization and mapping using particle
  filters.
\newblock {\em The International Journal of Robotics Research},
  25(12):1243--1256, 2006.

\bibitem{vacuum_clear_robot}
Will Knight.
\newblock The roomba now sees and maps a home.
\newblock {\em MIT Technology Review}, 2015.

\bibitem{HungAlgo}
H.~W. Kuhn.
\newblock The hungarian method for the assignment problem.
\newblock {\em Naval Research Logistics Quarterly}, 2(1‐2):83--97, 1955.

\bibitem{Autonomous_vehicle_2}
H.~{Lategahn}, A.~{Geiger}, and B.~{Kitt}.
\newblock Visual slam for autonomous ground vehicles.
\newblock In {\em 2011 IEEE International Conference on Robotics and
  Automation}, pages 1732--1737, 2011.

\bibitem{liu2019accurate}
Shuoyuan Liu, Peng Guo, Lihui Feng, and Aiying Yang.
\newblock Accurate and robust monocular slam with omnidirectional cameras.
\newblock {\em Sensors}, 19(20):4494, 2019.

\bibitem{montemerlo2002fastslam}
Michael Montemerlo, Sebastian Thrun, Daphne Koller, and Ben Wegbreit.
\newblock Fastslam: A factored solution to the simultaneous localization and
  mapping problem.
\newblock In {\em In Proceedings of the AAAI National Conference on Artificial
  Intelligence}, pages 593--598. AAAI, 2002.

\bibitem{munguia2007monocular}
Rodrigo Munguia and Antoni Grau.
\newblock Monocular slam for visual odometry.
\newblock In {\em 2007 IEEE International Symposium on Intelligent Signal
  Processing}, pages 1--6. IEEE, 2007.

\bibitem{mur2015orb}
Raul Mur-Artal, Jose Maria~Martinez Montiel, and Juan~D Tardos.
\newblock Orb-slam: a versatile and accurate monocular slam system.
\newblock {\em IEEE transactions on robotics}, 31(5):1147--1163, 2015.

\bibitem{quan2019tightly}
Meixiang Quan, Songhao Piao, Minglang Tan, and Shi-Sheng Huang.
\newblock Tightly-coupled monocular visual-odometric slam using wheels and a
  mems gyroscope.
\newblock {\em IEEE Access}, 7:97374--97389, 2019.

\bibitem{ICRA:18:landmark}
Rattanachai Ramaithitima and Subhrajit Bhattacharya.
\newblock Landmark-based exploration with swarm of resource constrained robots.
\newblock In {\em Proceedings of IEEE International Conference on Robotics and
  Automation (ICRA)}, May 21-25 2018.

\bibitem{subgradient_method}
A.~{Tahbaz-Salehi} and A.~{Jadbabaie}.
\newblock Distributed coverage verification in sensor networks without location
  information.
\newblock {\em IEEE Transactions on Automatic Control}, 55(8):1837--1849, 2010.

\bibitem{thrun2005multi}
Sebastian Thrun and Yufeng Liu.
\newblock Multi-robot slam with sparse extended information filers.
\newblock {\em Robotics Research}, pages 254--266, 2005.

\bibitem{Autonomous_vehicle_1}
Junqiao Zhao, Yewei Huang, Xudong He, Shaoming Zhang, Chen Ye, Tiantian Feng,
  and Lu~Xiong.
\newblock Visual semantic landmark-based robust mapping and localization for
  autonomous indoor parking.
\newblock {\em Sensors}, 19(1), 2019.

\end{thebibliography}

\end{document}